%% file: acl2023.tex
\pdfoutput=1

\documentclass[11pt]{article}

\usepackage[]{ACL2023}

\usepackage{listings}
\usepackage{amsmath}
\usepackage{cuted}
\usepackage{amsfonts}
\usepackage{pifont}
\usepackage{times}
\usepackage{latexsym}

\usepackage{empheq}
\usepackage{bbm}
\usepackage{tabularx}
\usepackage{hyperref}
\usepackage{mathtools}

\usepackage[T1]{fontenc}

\usepackage[utf8]{inputenc}

\usepackage{microtype}

\usepackage{inconsolata}

\definecolor{myred}{HTML}{CC3333}
\definecolor{mygreen}{HTML}{7EA6E0}

\usepackage{kantlipsum}
\usepackage[breakable, theorems, skins]{tcolorbox}
\tcbset{enhanced}

\usepackage{colortbl}

\newif\ifcomments
\commentstrue
\ifcomments
\usepackage[normalem]{ulem}
\definecolor{shaobopurple}{rgb}{0.8,0.0,0.8}

\newcommand\shaobo[1]{\textcolor{shaobopurple}{\textsf{\scriptsize[\textbf{shaobo\@:} #1]}}} 
\newcommand\shaoboi[1]{\textcolor{shaobopurple}{#1}} 
\newcommand\shaobom[1]{\marginpar{\raggedright\tiny\textcolor{shaobopurple}{\textsf{{\bfseries Shaobo\@:} #1}}}} 
\newcommand\shaobos{\bgroup\markoverwith{\textcolor{shaobopurple}{\rule[.4ex]{2pt}{0.8pt}}}\ULon} 
\else
\newcommand\shaobo[1]{}
\newcommand\shaoboi[1]{\ignorespaces}
\newcommand\shaobom[1]{}
\newcommand\shaobos[1]{#1}
\fi

\usepackage{xspace}
\newcommand{\SmallHeading}[1]{\noindent\textbf{#1.}\quad}
\newcommand{\Figure}{Fig.}
\newcommand{\Table}{Tab.}
\newcommand{\Appendix}{Appendix\xspace}
\newcommand{\Equation}{Equation\xspace}
\newcommand{\Section}{Section\xspace}
\newcommand{\frate}{\textit{fallacy-rate}~}
\newcommand{\Frate}{\textit{Fallacy-rate}~}
\newcommand{\FRate}{\textit{Fallacy-Rate}~}

\newcommand{\wrate}{\textit{win-rate}~}
\newcommand{\Wrate}{\textit{Win-rate}~}
\newcommand{\WRate}{\textit{Win-Rate}~}
\newcommand{\ourMethod}{FIPO\xspace}

\title{A Logical Fallacy-Informed Framework for Argument Generation}

\author{Luca Mouchel, Debjit Paul, Shaobo Cui, \\ \textbf{Robert West}, \textbf{Antoine Bosselut}, \textbf{Boi Faltings} \\
  EPFL, Switzerland \\
  \texttt{\{firstname.lastname\}@epfl.ch} \\
  }

\usepackage{booktabs}

\begin{document}
\maketitle
\begin{abstract}


Despite the remarkable performance of large language models (LLMs), they still struggle with generating logically sound arguments, resulting in potential risks such as spreading misinformation. An important factor contributing to LLMs' suboptimal performance in generating coherent arguments is their oversight of logical fallacies.
To address this issue, we introduce fallacy-informed preference optimization (FIPO) that helps steer LLMs toward generating logically sound arguments. 
\ourMethod~includes a classification loss to capture the fine-grained information on fallacy types. 
Our results on argument generation tasks show that \ourMethod reduces the fallacy errors by up to 17.5\%.
Furthermore, our human evaluation results reveal that the quality of the arguments generated by our method significantly outperforms the fine-tuned baselines and other preference optimization methods, such as DPO. These findings highlight the importance of ensuring models are aware of logical fallacies for effective argument generation.\footnote{Our code and datasets are publicly available for research purposes at
\href{https://github.com/lucamouchel/Logical-Fallacies}{github.com/lucamouchel/Logical-Fallacies}}

\end{abstract}

\section{Introduction} \label{}

\input{introduction}

\section{Related Work} 
\input{relatedwork}

\section{Task Formulation}
\input{prob-formulation}

\section{Methodology}
\input{method}

\section{Experimental Setup} 
\input{experimentalsetting}

\section{Experimental Results} \label{sec:results}
\input{results}

\section{Conclusion} \label{sec:conclusion} 
\input{conclusion}
\section*{Limitations}
Although various preference optimization strategies have shown improvement over the SFT baseline in reducing fallacious arguments, the margin remains modest. This may be attributed to several factors: our assumption that the original dataset \cite{cckg} was free of fallacies, the inherent complexity and diversity of fallacies which complicates effective detection, and the variability in model performance, particularly the weaker results from the {Mistral} model compared to {Llama-2}. Additionally, the limited size of our dataset and the brevity of arguments present further challenges, as the lack of contextual cues can hinder the models' ability to identify and avoid fallacies consistently.



\section*{Ethics Statement}
In this paper, we experiment with well-acknowledged datasets. Our framework improves argument generation in LLMs, which may encode biases related to race, gender, and other attributes \cite{ethics1,ethics2}. As our work does not mitigate these biases, the models may still reflect harmful behaviors. We recommend users deploying our model \textit{off-the-shelf} evaluate potential harm to protected groups and apply appropriate mitigation.
While improving argument generation is valuable, it poses risks if misused. Bad actors could exploit these capabilities to amplify disinformation, manipulate public opinion, or influence democratic processes by spreading persuasive yet harmful narratives. To address this, robust safeguards, such as usage policies, monitoring, and detection tools, are critical. Finally, our annotation task relied on AMT workers evaluating model-generated arguments, particularly for logical fallacies, which are complex to assess. Workers were English-speaking and paid adequately for their time to ensure fairness.

\section*{Acknowledgment}

We acknowledge the support of the ICT-48 Network of AI Research Excellence Center “TAILOR” (EU Horizon 2020, GA No 952215). Antoine Bosselut gratefully acknowledges the support of the Swiss National Science Foundation (No. 215390), Innosuisse (PFFS-21-29), the EPFL Center for Imaging, Sony Group Corporation, and the Allen Institute for AI.

\bibliography{anthology,custom}

\appendix
\onecolumn
\clearpage 
\section{Data Augmentation and Evaluation with LLMs}
\subsection{Generating Arguments with ChatGPT}
\label{gpt-fallacies}

Our prompt design for ChatGPT to generate fallacies follows a similar heuristic to \cite{liu-etal-2023-g}, by introducing the task, defining the fallacy type it must generate, along with two examples of that particular fallacy type. 
Following the distribution in \Figure~\ref{fig:distribution}, we generate four fallacies for the same topic and feed the following prompt to ChatGPT to generate fallacies as negative preference data:
\begin{center}
\begin{promptBox}{
You are given a topic \textit{T}.  
Your task is to generate a \{'supporting' or 'counter'\} 
argument in the form of a \textit{f-type}\footnote{Fallacy type that can be any of the thirteen types described in Table \ref{types_fallacies}} logical fallacy in the context of the topic. 
It should not be longer than 25 words. 

\textit{f-type} fallacy is defined as: \{definition\} \\
examples of \textit{f-type} are: \\
\{example 1\}\\
\{example 2\}

Here is an example of \textit{f-type} fallacy argument:  \\
\{example of an argumentative fallacy\}

return \{\\"topic": \textit{T}, "fallacy":~\textit{f-type}, "argument": <...>\\ \}
}
\end{promptBox}
\end{center}
Some examples of generated logical fallacies include: \textit{"I know someone who smoked cannabis and became successful. Therefore, everyone who smokes cannabis will be successful.", "I know a few people who spend too much time on social media and have no real-life friends. Therefore, social media is terrible for society."}. \Table~\ref{table:dpo_data} presents examples of samples in our preference dataset. Our augmented fallacy argument dataset consists of the train-test split in \Table~\ref{tab:traintestsplit}.

\subsection{Prompting GPT-4 for Evaluation}
\label{evalprompt}
To evaluate whether $\pi_\theta$ for a given method $\theta$ generates logical arguments compared to $\pi_\textnormal{SFT}$, we use GPT-4 and evaluate the \wrate (e.g., how often does $\pi_\theta$ produce better arguments) by prompting GPT-4 with: \begin{center}
\begin{promptBox}
{
        Which of these arguments is better for the topic {$t$} and stance $s$:\\
        1. $\pi_\textnormal{SFT}(y|t,s)$\\
        2. $\pi_\theta(y|t,s)$\\
        If both arguments are equally good, return 3 (Tie).
        The better argument is: <response>
}
\end{promptBox}
\end{center}

We also evaluate how often models produce logical fallacies, which we call \textit{fallacy-rate}, by prompting ChatGPT with:

\begin{center}
    \begin{promptBox}
       {
Consider the following $topic$ $t$, stance $s$ \{supporting or counter\} and argument $a=\pi_\theta(y|t,s)$:\\
Topic: $t$ \\
Argument: $a$\\
Out of all the following logical fallacy types \{list of types from \Table~\ref{types_fallacies}\} \\
would you qualify $a$ as one of these logical fallacies? If not - return "None".\\
If yes, which logical fallacy type is it? Let \textit{f-type}  be your answer. Return\\
\{"topic": $t$, "text": $a$, "fallacy type": \textit{f-type}
\}
} 
    \end{promptBox}
\end{center}

\section{Preference Optimization} \label{sec:literature:preference}
Preference optimization is a crucial step in aligning language models to generate outputs that meet user preferences and objectives effectively. It involves the process of ensuring that the goals and preferences of AI systems align with those of their human users. To demonstrate models are capable of learning to distinguish logically sound text from logical fallacies, we assess the performance of four different preference optimization techniques, including PPO \cite{ppo}, DPO \cite{DPO},
CPO \cite{cpo} and KTO \cite{kto}. 

\paragraph{PPO.} One widely used reinforcement learning optimization algorithm within RLHF is Proximal Policy Optimization (PPO). PPO \cite{ppo} is particularly favored due to its stability and efficiency. It iteratively updates the model's policy parameters by maximizing the expected cumulative reward while constraining the policy updates to a proximity threshold, preventing large policy changes that could destabilize learning.

\paragraph{DPO.}More recently, \citet{DPO} introduced Direct Preference Optimization, which skips the reward modelling part that is necessary for PPO. DPO leverages an analytical mapping from reward functions to optimal policies, to transform a loss function over reward functions into a loss function over policies and avoids fitting a reward model, while still optimizing under given preferences. 

\paragraph{CPO.}Another recently introduced method is Contrastive Preference Optimization (CPO). \citet{cpo} introduced CPO as a derivation of DPO, to address some shortcomings of DPO, including memory and speed inefficiencies. \cite{cpo} focuses mainly on machine translation, but the method can also be adapted to regular preference optimization for other tasks. They also incorporate a behaviour cloning regularizer to ensure that the policy does not deviate from the preferred data distribution. 

\paragraph{KTO.}Finally, the last method we evaluate is Kahneman-Tversky Optimization (KTO) \cite{kto}. The authors introduce the concept of human-aware loss functions (HALOs), which implicitly model human biases and have been shown to perform better than non-HALOs. Their approach directly maximizes the utility of generations instead of maximizing preference likelihood, as is commonly done.

\section{Details of Experimental Setup} \label{appendix:detailed_setup}
\subsection{Hyperparameters}
\label{appendix:detailed_setup:hyperparameters}
\SmallHeading{Training Parameters}For both base models, we train $\pi_\text{SFT}$ and  $\pi_\theta$ on 2 A100 GPUs for 3 epochs, using an Adam optimizer, with a learning rate of $2^{-4}$. 
For the LoRA configuration, we select a rank $r=16$, $\alpha=32$ and a dropout of $5^{-2}$. 
Regarding the specific training details of each method, we use $\beta=0.25$ for DPO \cite{DPO}, which controls how much  $\pi_\theta$ deviates from the reference model $\pi_\text{SFT}$. 
For both CPO \cite{cpo} and KTO \cite{kto}, we use $\beta=0.1$, which controls the implicit reward. Regarding the reward model for PPO \cite{ppo}, we use a binary logical fallacy classifier we trained using the chosen responses as non-fallacies and the rejected responses as fallacies. The classifier achieves accuracy and F1 over 95\% in the detection of fallacy arguments, which makes us confident in using the model as a reward model for this particular task and we use the logits as rewards. 

\SmallHeading{Decoding Parameters}
At inference time, we generate arguments with the aligned models using the same decoding strategy. We use nucleus-sampling with $p=0.75$ and top-$k$ sampling with $k=10$.

\SmallHeading{Hyperparameter Selection for \ourMethod} 
\label{appendix:lambda_tuning}
To optimize our custom loss function, defined in \Equation~\ref{custom loss function equation}, we conduct a series of experiments manipulating the hyperparameter $\lambda$ (\Equation~\ref{custom loss function equation}) and the weights for the cross-entropy loss (\Equation~\ref{eq:clf_loss}). 
Our initial step involved tuning the weights for the loss function. Given that our dataset consists of $n$ pairs of preferred and dispreferred arguments as logical fallacies, it is crucial to differentiate different fallacy classes. 
It is thus more suitable to set the weights for each fallacy type as their frequency in the dataset, given by \Equation~\ref{eq:clf_loss} and the weight for the preferred responses as little as possible, which we set as the minimum of all the fallacy frequencies. 
We also evaluate different settings of $\lambda$, testing values of 0.1, 0.3, and 0.6. Our findings indicate that a higher $\lambda$ effectively reduced the number of fallacies produced by the policies but adversely impacted the argument quality (\wrate). Conversely, a $\lambda$ of 0.1 had minimal impact on improving the \textit{\frate}. After assessing the trade-offs, we determined that a $\lambda$ value of 0.3 provided the optimal balance between minimizing fallacies and maintaining a reasonable \wrate.

\subsection{Context for Retrieval Augmented Generation (RAG)}
\label{appendix:detailed_setup:rag}
We use the \texttt{wiki-dpr} database to add contextual information to prompts. RAG~\cite{rag} ensures that arguments are grounded in factual accuracy and contextually relevant information by retrieving relevant information from pre-existing knowledge sources.
We retrieve relevant documents with the topics and add them to the prompts as context. For example, given the topic \textit{Factory farming should not be banned}, 
the knowledge extracted includes 
    \textit{"The practice of dairy production in a factory farm environment has been criticized by animal welfare activists. The U.S. Food and Drug Administration states that no 'significant difference' has been found between milk from treated and non-treated cows. [...]"}

\subsection{Licenses of Artifacts} \label{appendix:detailed_setup:licenses}
The artifacts we employ, including datasets, packages, and models, are detailed in \Table~\ref{tab:artifacts}. Our usage of these artifacts is consistent with their intended purposes.

\subsection{Detailed Setup of Human Annotations in Amazon Mechanical Turk} \label{appendix:human_annotation}
\begin{figure*}[!h]
    \centering
    \includegraphics[width=0.75\linewidth]{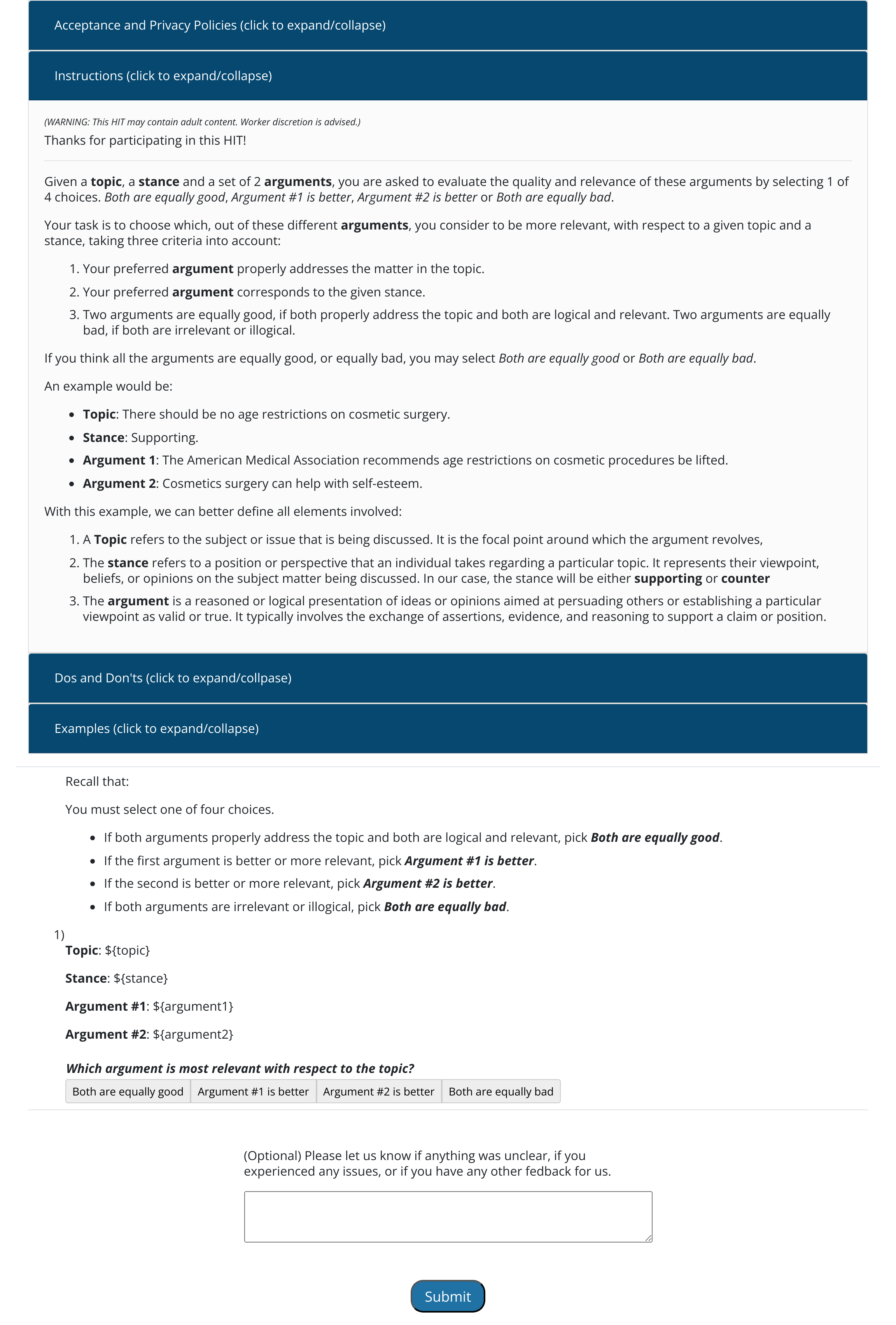}
    \caption{User interface used for human annotation on Amazon Mechanical Turk, where annotators compare pairs of arguments generated by different models to determine the superior argument.}
    \label{fig:comparative_amt_page}
\end{figure*}

The privacy of these collected annotations for the pairwise comparison is under the university policy, and we use Amazon Mechanical Turk's services. 
The webpage we use for the comparative study by human annotators is provided in \Figure~\ref{fig:comparative_amt_page}. The workers are paid fairly for their annotating work.

\section{Additional Experimental Results}

\subsection{Detailed \textit{fallacy-rates} in Zero-Shot Setting}
\label{appendix-zeroshot}
\Table~\ref{tab:zero-shot-fallacies} presents the distribution of fallacies detected with GPT-4 in the zero-shot setting using Llama-2 and Mistral in scenario \textbf{S$\mathbf{_1}$}: Prompting the models without any explicit instruction using logical fallacies-Simply providing a topic and stance and asking the model to generate an argument -- and scenario \textbf{S$\mathbf{_2}$}: including a definition of logical fallacy in the prompt, as well as two examples of logical fallacies and a clear instruction to not generate a logical fallacy.
\begin{table}[!h]
    \centering
    \resizebox{\linewidth}{!}{
    \begin{tabular}{lcccccccc}
        \toprule
        \textbf{Fallacy Types}  & \multicolumn{1}{>{\columncolor{red!20}}c}{Llama-2} & \multicolumn{1}{>{\columncolor{red!20}}c}{Mistral} & \multicolumn{1}{>{\columncolor{red!20}}c}{ChatGPT} & \multicolumn{1}{>{\columncolor{red!20}}c}{Llama-2-RAG} &\multicolumn{1}{>{\columncolor{blue!20}}c}{Llama-2} & \multicolumn{1}{>{\columncolor{blue!20}}c}{Mistral} &\multicolumn{1}{>{\columncolor{blue!20}}c}{ChatGPT}& \multicolumn{1}{>{\columncolor{blue!20}}c}{Llama-2-RAG}\\
        \midrule 
        Faulty Generalization      & 13 & 10 & 9 &12 & 9& 8& 6& 8 \\ 
        False Causality            & 3 & 4 & 4 & 5 & 4 & 3 & 3& 3 \\ 
        Appeal To Emotion          & 2 & 1 &1 & 2 & 1 & 2 &1 & 1\\
        Fallacy of Relevance       & 1 & 1 & - & 3 & 2 & 1& 2& 1\\ 
        False Dilemma              & 1& 1 & 2 & 3 & - & - & 1&1\\ 
        Circular Reasoning         & 29 & 21 &3 & 12 & 5 & 2 &- & 5\\ 
        Fallacy of Logic           & 6 & - & 2 & - & - & 2 & 1&-\\
        \midrule
        \FRate              & 55 & 38 & 21 & 37 & 21 & 18 & 14 & 19\\
        Not A Fallacy              & 45 & 62 & 79& 63 & 79 & 82 & 86 & 81\\
        \bottomrule \\
    \end{tabular}
    }
    \caption{\Frate according to GPT-4 on arguments generated by Llama-2,  Mistral, ChatGPT, and a Llama-2 based RAG model on scenario \textbf{S$\mathbf{_1}$} ( \colorbox{red!20}{\makebox[0.5cm]{\rule{0pt}{0.15cm}}} ) and scenario \textbf{S$\mathbf{_2}$} ( \colorbox{blue!20}{\makebox[0.5cm]{\rule{0pt}{0.15cm}}} ).}
    \label{tab:zero-shot-fallacies}
\end{table}
\vspace{-3mm}
\subsection{Results Regarding \WRate}
\label{appendix-automated-win-rates}
\Table~\ref{tab:model-win-rates} presents the results of win-rates computed by GPT-4 by comparing the policies with the corresponding SFT models and comparing which models win more often.

\begin{table}[htp!]
    \centering
    \begin{tabular}{lccc|lccc}
        \toprule
        \textbf{\WRate}  & $\pi_\textbf{SFT}$ & $\pi_\theta$ & \textbf{Tie} & $\pi_\textbf{SFT}$ & $\pi_\theta$ & \textbf{Tie}\\
        \midrule
        &\multicolumn{3}{>{\columncolor{gray!20}}c}{\textit{Llama-2 (7B)}} & \multicolumn{3}{>{\columncolor{gray!20}}c}{\textit{Mistral (7B)}}\\ 
        $\pi_\textbf{SFT}$ vs. $\pi_\textbf{DPO}$     & 35 & 61 & 4  & 49 & 48&3  \\ 
        $\pi_\textbf{SFT}$ vs. $\pi_\textbf{PPO}$&38&49&13 &48&49&3 \\
        $\pi_\textbf{SFT}$ vs. $\pi_\textbf{CPO}$& 43.5& 52.5 &4& 46.75& 51.5&1.75\\
        $\pi_\textbf{SFT}$ vs. $\pi_\textbf{KTO}$&40.5& 46& 13.5&44.5& 51.5& 4\\
        $\pi_\textbf{SFT}$ vs. $\pi_\textbf{FIPO}$&33& \textbf{63.5}& 3.5&27& \textbf{68}& 5\\
        \bottomrule \\
    \end{tabular}
    \caption{\Wrate according to GPT-4. $\pi_\theta$ is the aligned policy, where $\theta$ designates the respective alignment method (e.g., when comparing $\pi_\textbf{SFT}$ with $\pi_\textbf{DPO}$, then the value 61 means $\pi_\textbf{DPO}$ wins 61\% of the time.}
    \label{tab:model-win-rates}
\end{table}

Another interesting observation from \Figure~\ref{fig:manual-win-rate-annotations} and \Table~\ref{tab:model-win-rates} is the tie-rate.
Annotating concise arguments to determine superiority is challenging for humans, often leading to the simpler selection of \textit{Tie} when both arguments adequately address the topic. In fact, \textit{Tie} was chosen more frequently in human evaluations compared to the automatic evaluation by GPT-4.

\subsection{Examples of Fallacies Produced during Inference}
\label{appendix:argument examples}
\begin{figure*}[!h]
    \centering
    \includegraphics[width=1\linewidth]{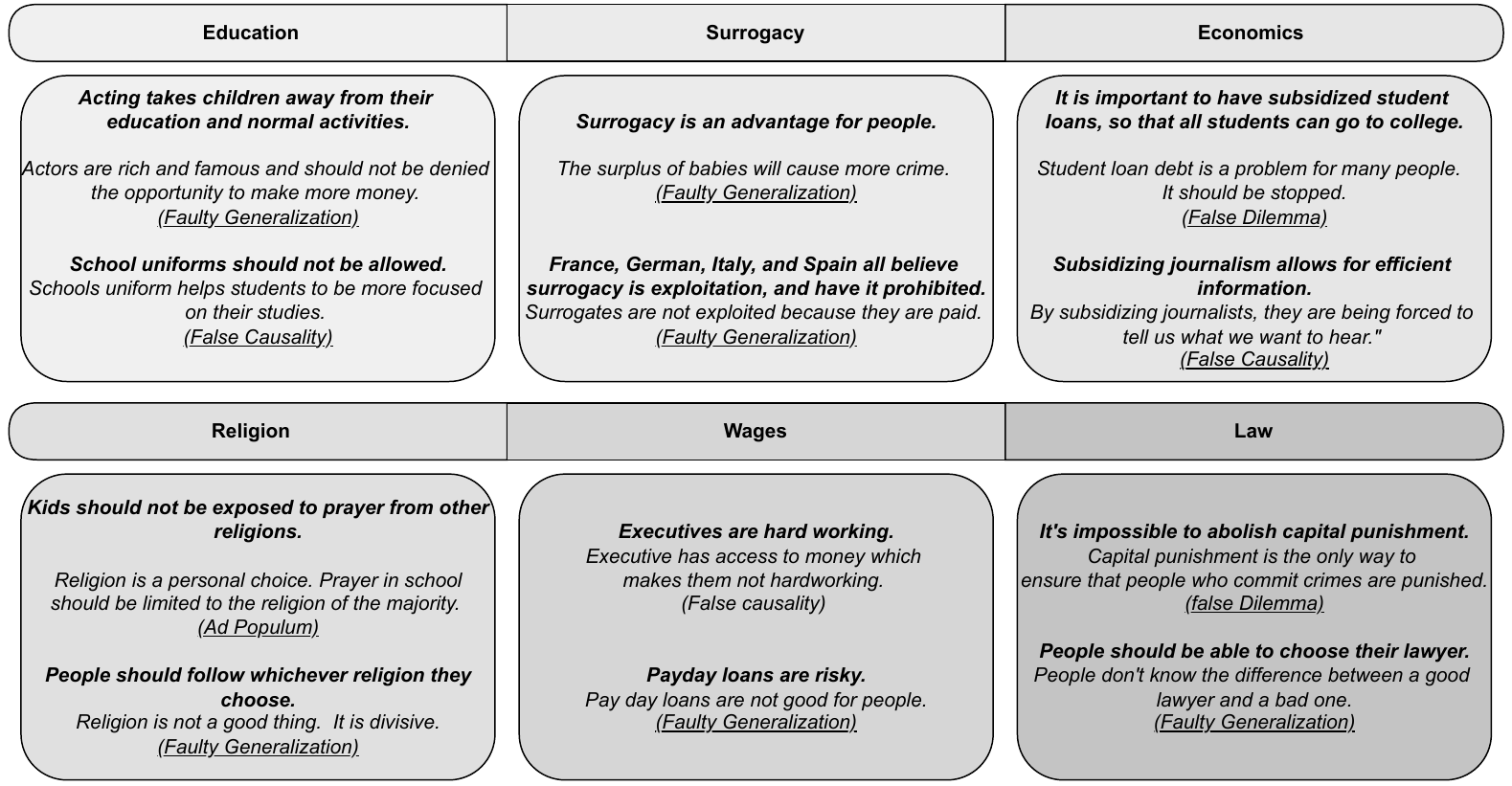}
    \caption{Examples of fallacious arguments generated at inference time by different models. }
    \label{tab:examples_fallacies}
\end{figure*}

\Table~\ref{tab:examples_fallacies} shows a few examples of arguments and fallacies produced by the models on a set of three topics.
\section{Using Humans to Classify Arguments}
\label{appendix:human_classification}
We use annotators to compute agreement scores with GPT-4's predictions, in order to justify the use of GPT-4 and its reliability. Another annotation task was conducted prior, and involved annotators tasked with classifying arguments into one of 13 fallacy types or as \textit{Not a Fallacy}. Out Of 200 samples, each annotated by three different workers, a majority agreement was reached in only 47\% of cases. This is likely due to the inherent difficulty of accurately identifying logical fallacies and leads to a loss of information if no agreement is reached. As such, we believe the easiest way to involve humans in computing the \frate is to task them with a binary decision instead.

\section{Fallacy-Informed Preference Optimization: Loss Definition}
\label{appendix:loss definition}
In our work, FIPO combines the CPO loss \cite{cpo} for preference optimization with our classification loss. We selected CPO because it achieves the optimal balance between \wrate and \frate (\Figure~\ref{fig:manual-win-rate-annotations}, \Table~\ref{tab:fallacies-gpt4}). Moreover, the CPO approach is designed to make the model learn rejections more accurately, which fits our study.\\

\noindent\textbf{CPO Loss} \hspace{0.6cm}
CPO is a reference-free preference optimization method. 
It extends DPO \cite{DPO}, modifying the DPO loss by using a Uniform model instead of a reference model.
The DPO loss is defined as:
$$
    \mathcal{L}_\text{DPO}(\pi_\theta, \pi_\text{ref}) = - \mathbb{E}_{(x,y_w,y_l)\sim \mathcal{D}} \left[ \log\sigma\left(\beta\log\frac{\pi_\theta(y_w|x)}{\pi_\text{ref}(y_w|x)} -\beta\log\frac{\pi_\theta(y_l|x)}{\pi_\text{ref}(y_l|x)}
    \right)
    \right]
$$
The CPO loss first approximates the DPO loss using a uniform reference model:
$$
    \mathcal{L}(\pi_\theta, U) = - \mathbb{E}_{(x,y_w,y_l)\sim \mathcal{D}} \Big[ \log\sigma\Big(\beta\log\pi_\theta(y_w|x) -\beta\log\pi_\theta(y_l|x)\Big)
    \Big]
$$
and defines the CPO loss as $
    \mathcal{L}_\text{CPO}(\pi_\theta)=\min_\theta \Big(\mathcal{L}(\pi_\theta, U)-\mathbb{E}_{(x,y_w)\sim \mathcal{D}}[\log \pi_\theta(y_w|x)]\Big)
$
which combines the preference loss with a negative log-likelihood (NLL) term. The $\min$ term ensures $\pi_\theta$ does not deviate from the preferred data distribution.\\

\noindent\textbf{FIPO Loss}\hspace{0.6cm}
Recall our preference dataset $\mathcal{D}=\{
t^{(i)},
s^{(i)},
y_w^{(i)},
y_l^{(i)},
k^{(i)}
\}$, with $t$ the topic, $s$ the stance, $y_w$ the preferred valid argument, $y_l$ the dispreferred fallacious argument and $k$ the fallacy type of $y_l$.

\noindent Now recall the definitions for our classification loss in~Section \ref{sec:loss_func}. We define the probability $p_k$ of each fallacy type $k$ as the output of the linear layer after we feed it the last hidden state after the forward pass into the model $\pi_\theta$ for both the preferred and dispreferred samples. That is, we perform a forward pass with $y_w$ and $y_l$ and obtain the last hidden states from the last tokens: ${\mathbf{h}_\theta}(y_{\{w,l\}}|t,s) = \pi_\theta^{L,T}(y_{\{w,l\}}|t,s)$ where $L, T$ are the number of layers in the language model and the position of the last token respectively. From the hidden states, we compute the probabilities using:
\begin{equation}
    \mathbb{P}_{\mathbf{h}_\theta}^k(y_{\{w,l\}}|t,s) = \text{Softmax}(\mathbf{Wh}_\theta(y_{\{w,l\}}|t,s) + \mathbf{b})_k; \text{ with Softmax}(z_i) = \frac{e^{z_i}}{\sum^K_{j=1}e^{z_j}}
\end{equation} 
where $W$ is the linear layer's weight matrix, $b$ is the corresponding bias term.
This way, we can compute the log probabilities for the preferred and dispreferred samples $\psi_0(y_w|t,s) = \log \mathbb{P}^0_{\mathbf{h}_\theta}(y_w|t,s)$ and  $\psi_k(y_l|t,s)= \log \mathbb{P}^k_{\mathbf{h}_\theta}(y_l|t,s)$ respectively. 
With these $\log$ probabilities, we can compute a cross-entropy loss for both samples:
\begin{equation}
    \begin{cases}
        \mathcal{L}_w &= -\mathbb{E}_{(t,s, y_w, y_l, k)} \left[\psi_0(y_w|t,s)\right]\\
        \mathcal{L}_l &= -\mathbb{E}_{(t,s, y_w, y_l, k)} \left[\psi_k(y_l|t,s)\right]
    \end{cases}
\end{equation}

\noindent Additionally, we use weights in our classification loss defined as the frequency of the fallacy types in our dataset to guide the model towards generating logical arguments by penalising fallacy errors more.
$$
     w_k = \frac{1}{|\mathcal{D}|}\sum\limits_{i = 1}^{\vert \mathcal{D}\vert}\mathbbm{1}\{{k^{(i)}=k}\} \hspace{0.2cm} \text{For \textit{fallacy-types }} k; \hspace{0.2cm}
     w_0 = \min\limits_k w_k \hspace{0.2cm} \text{For the class \textit{'Not a Fallacy'} (class 0) }
$$
Our method introduces a weighted cross-entropy loss which we use in addition to the preference loss:
\begin{align}
    \mathcal{L}_\text{CLF} &= w_0 \mathcal{L}_w + w_k \mathcal{L}_l
    \\ 
    &=-w_0\mathbb{E}_{(t,s, y_w, y_l, k)\sim \mathcal{D}}
    \Big[\psi_0(y_w|t,s)\Big]  -w_k\mathbb{E}_{(t,s, y_w, y_l, k)\sim \mathcal{D}} \Big[\psi_k(y_l|t,s)\Big]\\
    &=-\mathbb{E}_{(t,s, y_w, y_l, k)\sim \mathcal{D}}
    \Big[w_0\log \mathbb{P}^0_{\mathbf{h}_\theta}(y_w|t,s) + w_k\log \mathbb{P}^k_{\mathbf{h}_\theta}(y_l|t,s)\Big]
\end{align}
FIPO is thus reference-free and the loss $\mathcal{L}_\text{FIPO}$ is then defined with the following term:
\begin{lucasmall}
\begin{align*}
   \mathcal{L}_\text{FIPO} &= \mathcal{L}_\text{CPO} + \lambda\mathcal{L}_\text{CLF}\\
   &= -\mathbb{E}_{(t,s,y_w, y_l, k)\sim \mathcal{D}} 
   \Big[
   \underbrace{
   \min_\theta 
     {\log\sigma\Big(\beta\log\pi_\theta(y_w|t,s) -\beta\log\pi_\theta(y_l|t,s)\Big)}+
    {\log \pi_\theta(y_w|t,s)}}_{\mathcal{L}_{\text{CPO term}}} \\
    &\;\;\;\;\;\;+ \lambda\underbrace{\Big(
    w_0\log \mathbb{P}^0_{\mathbf{h}_\theta}(y_w|t,s) + w_k \log \mathbb{P}^k_{\mathbf{h}_\theta}(y_l|t,s) \Big)}_{{
    \mathcal{L}_{\text{CLF term}}}}
    \Big] 
\end{align*}
\end{lucasmall}

\begin{table*}[htbp]
\newcommand{\graycline}[0]{\arrayrulecolor{gray!70}\cline{0-2}\arrayrulecolor{black}}
\centering
\begin{tabularx}{\textwidth}{p{3.09cm}p{3.7cm}X}
\toprule
 \textbf{Term} & \textbf{Alternate Term~(if any)} & \textbf{Definition} \\
\hline
 Faulty Generalization & N/A & When a conclusion is drawn about all or many instances of a phenomenon. It is an example of jumping to conclusions. \\
\graycline
 False Causality & N/A & Statement that jumps to a conclusion implying a causal relationship without supporting evidence. \\
\graycline
 Circular Reasoning & N/A & when the end of an argument comes back to the beginning without having proven itself. \\
\graycline
 Ad Populum & Appeal to Popularity & an argument based on affirming that something is real or better because the majority thinks so. \\
\graycline
Ad Hominem & Personal Attack & Irrelevantly attack the person or some aspect of the person making the argument. \\
\graycline
 Fallacy of Logic & Logical Error & An error in the logical structure of an argument. \\
\graycline
 Appeal to Emotion & N/A & Manipulation of the recipient's emotions in order to win an argument \\
\graycline
 False Dilemma & Excluding Viable Alternatives & Presenting only two options or sides when there are many options or sides. \\
\graycline
Equivocation & Use of Ambiguous Language & When a key term or phrase in an argument is used in an ambiguous way. \\
\graycline
 Fallacy of Extension & Exaggerating & Attacking an exaggerated or caricatured version of your opponent's position. \\
\graycline
 Fallacy of Relevance & Irrelevant Argument & Introducing premises or conclusions that have nothing to do with the subject matter. \\
\graycline
Fallacy of Credibility & Attacking the Character of the Speaker & Attempts to disprove an argument by attacking the character of the speaker. \\
\graycline
Intentional & Intentionally Wrong Argument & Some intentional (sometimes subconscious) action/choice to incorrectly support an argument \\
\bottomrule
\end{tabularx}
\caption{Definitions of different types of logical fallacies. }
\label{types_fallacies}
\end{table*}

\begin{table*}[htbp]
\centering
\resizebox{\linewidth}{!}{
\begin{tabular}{p{6.2cm}p{6.2cm}p{6.2cm}}
\toprule
\textbf{Prompt} &  \textbf{Golden} & \textbf{Fallacy} \\ \hline
Generate a supporting argument for the topic: Cannabis should be legal. & It's not a bad thing to make marijuana more available. & Why should we be worrying about legalizing cannabis when there are more important issues like poverty and hunger? (\textit{Fallacy of Relevance})\\ \hline
Generate a supporting argument for the topic: Urbanization is terrible for the planet. & Urbanization increases pollution.& Either we continue urbanization and destroy the planet, or we stop urbanization and hinder economic growth. (\textit{False Dilemma})\\ \hline
Generate a supporting argument for the topic: Research on embryonic stem cell should not be tax subsidized because for many it goes against their religious beliefs. & There are Christians who disagree with doing research on embryonic stem cells.& Those who support tax subsidies for embryonic stem cell research are godless and immoral. (\textit{Ad Hominem}) \\ 
\bottomrule
\end{tabular}
}
\caption{Examples of samples from the preference dataset used for preference optimization. Golden arguments are retrieved from previous work \cite{cckg}, while fallacy arguments are generated using ChatGPT.} 
\label{table:dpo_data}  
\end{table*}

\begin{table*} 
    \centering
    \resizebox{\textwidth}{!}{
    \begin{tabular} 
    {llll} 
       \toprule 
        \textbf{Artifacts} & \textbf{Citation} & \textbf{Link} & \textbf{License}\\ 
        \midrule
        \textsc{Logic} &  \cite{LOGIC} & \url{https://github.com/causalNLP/logical-fallacy}  & N/A \\
        \textsc{ExplaGraphs} & \cite{cckg} & \url{https://explagraphs.github.io/}  & CC BY-SA 4.0 \\
        Debatepedia & \cite{debatepedia} & N/A  & N/A \\
        PyTorch & \cite{paszke-etal-2019-pytorch} & \url{https://pytorch.org/} & BSD-3 License\\
        transformers & \cite{wolf-etal-2020-transformers} & \url{https://huggingface.co/docs/transformers/index} & Apache License 2.0\\
        wandb & N/A & \url{https://pypi.org/project/wandb/} & MIT License\\
        nltk & \cite{bird-loper-2004-nltk} & \url{https://www.nltk.org/} & Apache License 2.0 \\
        OpenAI API & N/A & \url{https://platform.openai.com/docs/api-reference} & MIT License\\
        scikit-learn & \cite{pedregosa-etal-2011-scikit-learn} & \url{https://scikit-learn.org/stable/} & BSD License \\
      \bottomrule 
    \end{tabular}
    }
    \caption{Licenses of the artifacts utilized in this work, including datasets and major software packages, along with their respective licenses.  }
    \label{tab:artifacts}
\end{table*}

\clearpage

\end{document}

%% file: introduction.tex
Argument generation is crucial in daily life and has numerous online and offline applications. For instance, legislative bodies often use persuasive arguments to secure the necessary votes for bills to pass. However, generating logically coherent arguments is a challenging task and requires an appropriate combination of \textit{reliable evidence} and \textit{effective logical reasoning} \cite{walton2008argumentation, wachsmuth-etal-2017-argumentation}.
Humans are prone to misconstruing logical argumentation in the real world and often unknowingly adopt flawed reasoning in discussions \citep{humanreasoning}. Similarly, large language models (LLMs) have demonstrated limitations in their logical reasoning capabilities, suffering from logical inconsistencies \citep{chen2023learning, LOGIC,CBR}, and producing logically incorrect arguments \citep{Chen2023ExploringTP}. 

\begin{figure}[!t]
    \centering
    \includegraphics[width=\linewidth]{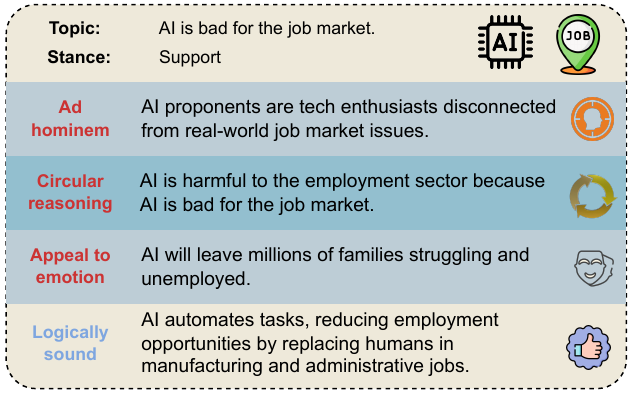}
    \caption{Examples of \textcolor{myred}{fallacious} and \textcolor{mygreen}{logically sound} arguments.}
    \label{fig:intro}
\end{figure}

In this work, we hypothesize that LLMs generate logically incorrect arguments because they lack an understanding of logical fallacies. A logical fallacy is an error in reasoning that undermines the validity of an argument \citep{tindale2007fallacies}. For example, \textit{"I’ve never had the flu because I take my vitamins every day."} is an instance of a \textit{false causality} fallacy. These fallacies arise from unsound premises. They can be identified by the absence of legitimate and relevant evidence to support their claims.

In a preliminary study, we evaluate $100$ arguments on different topics with ChatGPT and find that $21$\% of the arguments contain fallacies. 
We observe that several types of logical fallacy arguments, especially \textit{false causality and faulty generalization} are commonly generated by different LLMs such as Llama-2~\cite{touvron2023llama} or Mistral~\cite{mistral}. 
Our study explores the relationship between logical fallacy understanding and argument generation. We view models generating logically coherent arguments as a logical alignment problem, i.e., aligning the model responses (arguments) to the given topic and stance. Recent methods, such as Reinforcement Learning with Human or AI Feedback, have been shown to bridge the “alignment gap” between model responses and human preferences \cite{christiano2017deep_rl,ziegler2019fine,lee2023rlaif}. 
However, LLMs fine-tuned with RLHF can still generate logically fallacious arguments. Hence, to train models to prefer logically correct arguments, it is important to have reliable and diverse error scenarios as training examples. To address this issue, we define $13$ categories of logical fallacy errors, drawing inspirations from the history of logic and logical fallacies studied since the times of Ancient Greece by Aristotle  \citep{aristotle}. Figure \ref{fig:intro} depicts some fallacy examples, and Figure \ref{fig:distribution} shows different fallacy categories. 

We use ChatGPT to collect $7,872$ fallacy arguments spanning different fallacy categories to train preference models. 
First, we perform supervised fine-tuning (SFT) to teach models to generate arguments. Next, we use preference optimization methods to instil the ability to generate logically correct arguments. Specifically, we use Direct Preference Optimization (DPO) \citep{DPO}, Proximal Policy Optimization (PPO) \citep{ppo}, Kahneman-Tversky Optimization (KTO) \citep{kto}, and Contrastive Preference Optimization (CPO) \citep{cpo} on our preference dataset. 
These fallacy-informed models demonstrate a notable improvement in argument quality, achieving a higher \wrate (i.e., the proportion of wins over the SFT baseline in terms of argument quality) and reducing the \frate (i.e., the proportion of fallacies generated) by up to 8.5\%. However, we observe that the above methods fail to account for the nuanced differences between logical fallacies, instead treating each fallacy similarly. 

To this end, we introduce \textbf{F}allacy-\textbf{I}nformed \textbf{P}reference \textbf{O}ptimization (\ourMethod) that combines the original preference optimization loss with a weighted cross-entropy classification loss. This additional loss 
penalizes the model based on the frequency of different fallacies in the preference dataset, applying stronger penalties for misclassifying more occurring fallacies during training and reinforcing fallacy-aware learning.
We observe that \ourMethod outperforms the SFT baselines by reducing the \frate from 34.5\% to 17\% for Llama-2 (7B) and from 32.5\% to 19.5\% for Mistral (7B). 
FIPO also outperforms the best preference optimization method (PPO-Llama 2 and KTO-Mistral) by producing 9\% and 8.25\% fewer logical fallacy errors, respectively.
Our analysis explores how preference optimization enhances argument quality and compares FIPO to other methods. We also investigate whether preference optimization reduces logical fallacy errors and how FIPO improves upon existing techniques. Additionally, we examine the most common fallacy types observed in arguments generated during inference.

\noindent \textbf{Contributions:}
    (i) To the best of our knowledge, we are the first to study and show how understanding logical fallacies can improve argument generation quality; 
    (ii) We introduce \ourMethod, which integrates a classification loss during the preference optimization phase, which helps further reduce the \textit{fallacy-rate};
    (iii) Human evaluation results validate GPT-4's reliability in identifying fallacies and FIPO's higher-quality arguments compared to other preference learning methods.

%% file: relatedwork.tex

\begin{figure*}[t]
    \centering
    \includegraphics[width=\linewidth]{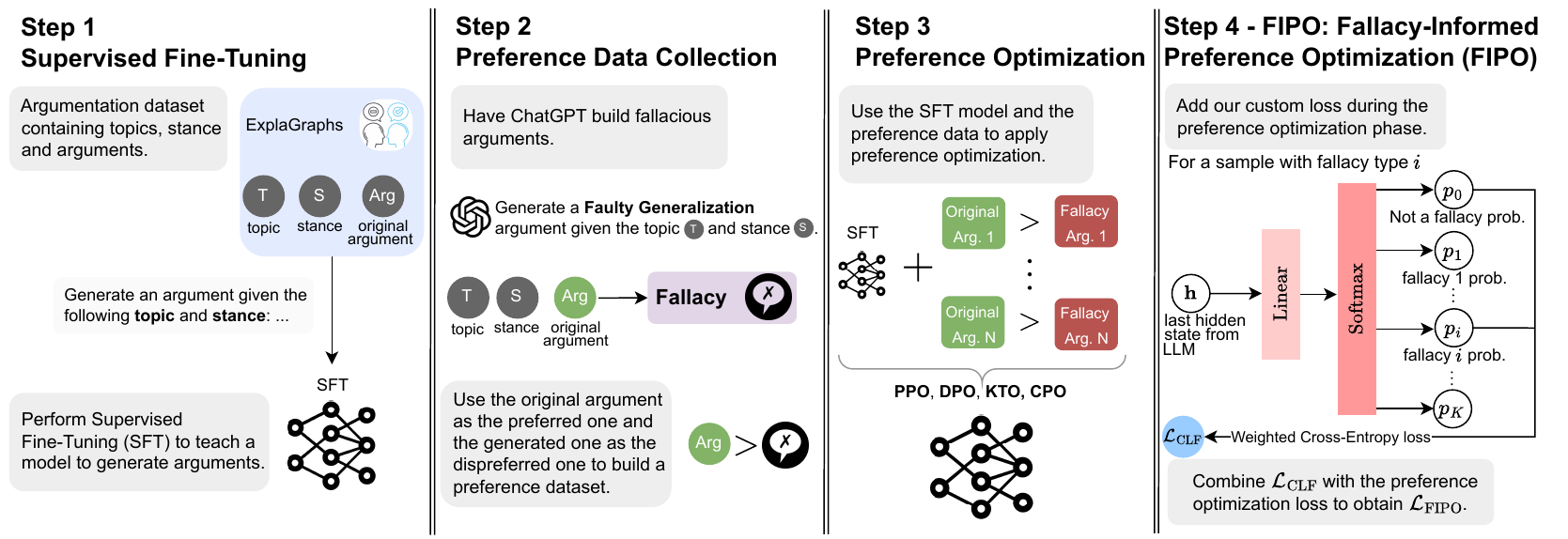}
    \caption{Overview of our framework. 
    The first step is supervised fine-tuning using argumentation data. Next, we collect preference data by generating fallacious arguments using ChatGPT. We then perform preference optimization using methods like DPO, PPO, CPO, and KTO. Finally, we introduce FIPO, which integrates a classification loss during the preference optimization phase. 
    }
    \label{fig:pipeline}
    \vspace{-5mm}
\end{figure*}
\paragraph{Logical Fallacies.}
{Logical fallacies are errors in reasoning that can undermine the validity of an argument \citep{tindale2007fallacies}. In argumentative discourse, identifying fallacies is crucial for measuring the quality of argumentation \citep{wachsmuth-etal-2017-argumentation, nakpih2020automated}. 
Ac Prior works have shown that LLMs struggle to classify logical fallacies, with F1 scores reaching 66\% \citep{LOGIC, CBR,ruiz-dolz-lawrence-2023-detecting}. 
}
More recently, \citet{li2024reason} demonstrated GPT-4's ability to identify and classify fallacies, achieving over 86\% accuracy in both tasks.
Nevertheless, previous works have not explored how a nuanced understanding of logical fallacies might influence argument generation. 





\paragraph{Argument Generation.}
Argument generation is an important task in natural language processing that involves generating coherent and persuasive arguments for a given topic. Existing argument generation frameworks have made significant strides: \citet{hua-wang-2018-neural} introduced a generator that creates arguments from key phrases, followed by a separate decoder to produce the final argument text. \citet{hua-etal-2019-argument-generation} developed Candela, a style-controlling counter-argument generation framework. \citet{schiller-etal-2021-aspect} presented Arg-CTRL, a model that uses control codes for topic, stance, and aspect in sentence-level argument generation. More recently, \citet{saha-srihari-2023-argu} introduced an argument generator for factual arguments across a limited set of topics. Despite these advances, no study has yet addressed generating arguments from the lens of logical fallacies.

\paragraph{Data Generation and Automatic Evaluation with LLMs.}
Using LLMs in data generation is supported by their proven effectiveness in a spectrum of text generation tasks, including the creation of instructional and relational datasets \cite{peng2023instruction,sun2023principledriven,wang-etal-2023-self-instruct,10.5555/3618408.3619681}. Notably, \citet{schick-schutze-2021-generating} demonstrated the utility of LLMs in producing datasets that significantly enhance the training of smaller models. 
Regarding the evaluation of automatically generated text, \citet{liu-etal-2023-g} highlight that traditional metrics such as BLEU~\cite{papineni-etal-2002-bleu} and ROUGE~\cite{lin-2004-rouge} are inadequate for tasks requiring creativity and diversity. Given \citet{li2024reason}'s demonstration of GPT-4's ability to identify and classify logical fallacies, we use it as a judge to identify fallacies in arguments--validated by a human annotation task we performed to verify GPT-4's reliability (\Section~\ref{gpt4-reliability}).

%% file: prob-formulation.tex
\label{pbformulation}
In this work, we address the argument generation task. In order to assess LLMs' capabilities for argument generation, we leverage the \textsc{ExplaGraphs} dataset \cite{cckg}, consisting of topics, stances and arguments, denoted as $\mathcal{D}\!=\!\{t^{(i)}, s^{(i)}, y_w^{(i)} \}^{N}_{i=1}$, where $t$ is the topic, $s$ the stance (supporting or counter), and $y_w$ the argument. 
One naive approach to address the problem of logical argument generation is prompting LLMs. To assess this approach, we evaluate ChatGPT (\texttt{gpt-3.5-turbo}), Llama-2 (7B), and Mistral (7B) in the zero-shot setting on a set of 100 topics. Additionally, we implement a Retrieval Augmented Generation (RAG) model with Llama-2 using the \texttt{wiki-dpr} database \cite{wiki-dpr}. Examples of contexts retrieved for RAG are provided in \Appendix~\ref{appendix:detailed_setup:rag}. 
This baseline evaluation is made on two separate scenarios: in \textbf{S$_\mathbf{1}$}, we prompt models to generate arguments given a topic and a stance. In \textbf{S$_\mathbf{2}$}, we guide the model towards generating logical arguments by defining logical fallacies, giving two examples, and instructing them not to generate a fallacious argument.
\textbf{We observe models struggle to generate logically sound arguments in S$\mathbf{_1}$.} In \Table~\ref{tab:challenging_task}, we report the performance of all the models in argument generation.  Since GPT-4 (\texttt{gpt-4} on OpenAI's API) is a good fallacy identifier \cite{li2024reason}, further validated by our own human annotation task to verify reliability (\Section~\ref{gpt4-reliability}), we use it to assess the \frate of the generated arguments. In \textbf{S$_\mathbf{1}$}, ChatGPT outperformed the open source models--however, it still generates fallacious arguments in 21\% of the cases. \textbf{We notice a very sharp improvement in S$\mathbf{_2}$}, implying explicit knowledge of fallacies and examples in prompts help generate logical arguments. While we include \textbf{S$_\mathbf{2}$} as a baseline to assess the impact of explicit guidance, our study examines LLMs' inherent ability to generate logical arguments without assistance—mirroring the conditions of \textbf{S$_\mathbf{1}$}.
The detailed distributions of fallacy types across different approaches in the zero-shot setting are presented in \Table~\ref{tab:zero-shot-fallacies}.
\begin{table}[!h]
    \centering
    \resizebox{\linewidth}{!}{
    \begin{tabular}{lcccc}
    \toprule
        Model & ChatGPT & Llama-2 & Mistral & Llama-2-RAG\\
    \midrule
        \frate \textbf{S$_\mathbf{1}$} & 21& 55 & 38& 37\\
        \frate \textbf{S$_\mathbf{2}$} & 14& 21 & 18 & 19\\  
    \bottomrule
    \end{tabular}
    }
    \caption{\frate for arguments generated by different baselines.}
    \label{tab:challenging_task}
\end{table}

%% file: method.tex
To address the challenge of generating fallacy-free arguments, we propose using preference learning methods to generate arguments logically aligned with the given topic and stance. 
This approach involves making models aware of logical fallacies and training them to generate logically correct arguments by rewarding valid arguments and penalizing dispreferred samples. 
The process of preference learning typically involves three main steps: (i) supervised fine-tuning (SFT) (\Section\ref{sec:methodology:sft}), (ii) preference data collection (\Section~\ref{sec:methodology:augmentation}) and (iii) reinforcement learning (\Section~\ref{sec:rl_phase}). In \Section~\ref{sec:loss_func}, we introduce our method (\ourMethod), which introduces fine-grained information about fallacies in the alignment process.  A comprehensive overview of the methodology is presented in \Figure~\ref{fig:pipeline}.
To justify our design and methodology for FIPO and the preference data collection, we perform two ablation studies with different training approaches, described in \Section~\ref{appendix:ablation study}. The results demonstrate that our design achieves the best performance.



\subsection{Supervised Fine-Tuning} \label{sec:methodology:sft}
We fine-tune a pretrained language model $\pi_\beta$ on the \textsc{ExplaGraphs} \cite{cckg} dataset $\mathcal{D}$ with maximum likelihood estimation to obtain $\pi_{\text{SFT}}$. 
\begin{equation}
\hspace{-2mm}
    \mathcal{L}_\text{SFT}(\pi_{\beta})\!=\! -{\mathbb{E}}_{(t,s,y_w) \sim \mathcal{D}}\big[\log \left(\pi_{\beta}(y_w|t, s)\right)\!\big]
\end{equation}

\subsection{Preference Data Collection} 
\label{sec:methodology:augmentation}
\begin{figure}[!h]
    \centering
    \includegraphics[width=\linewidth]{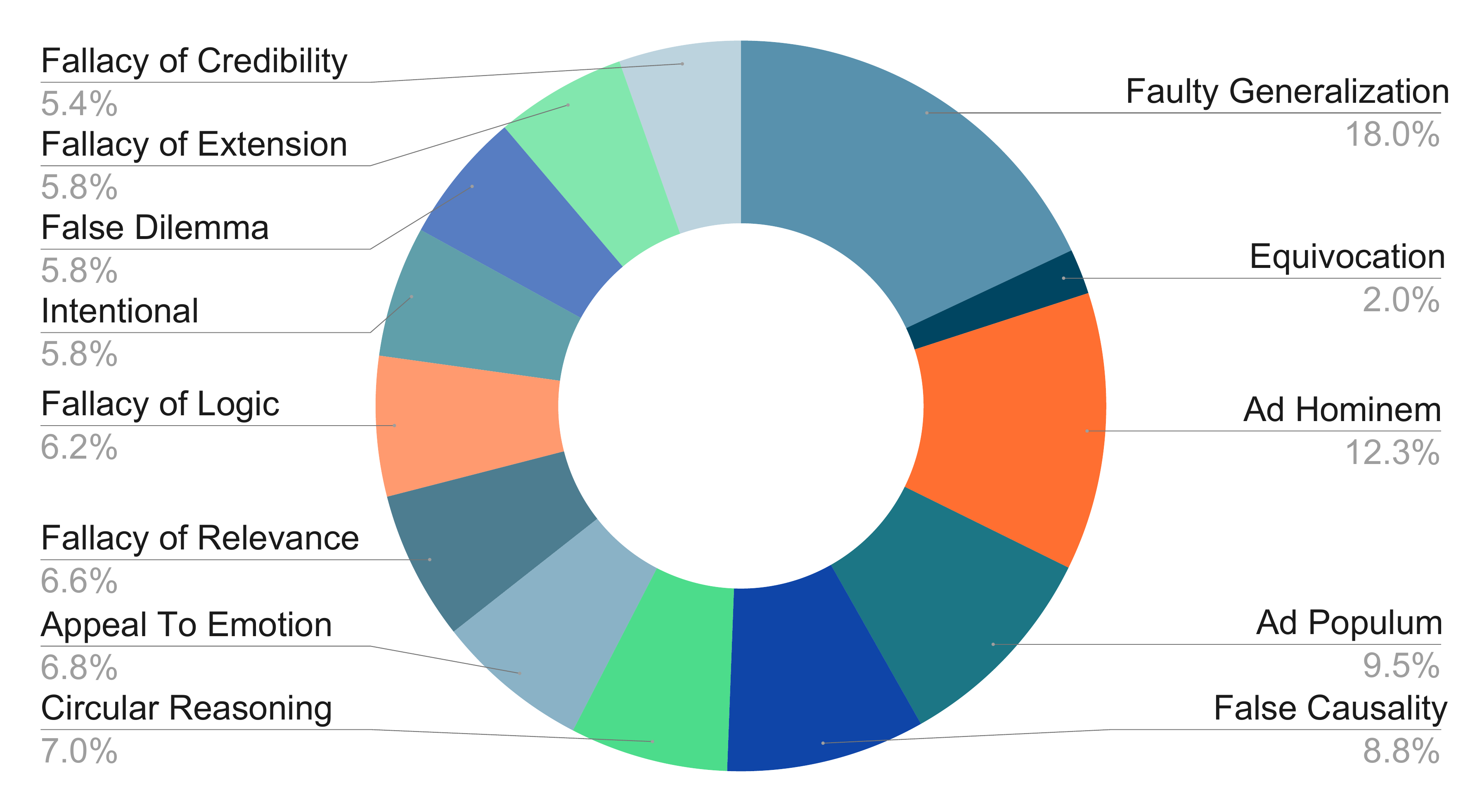}
    \caption{Distribution of different fallacy types according to the LOGIC dataset  \cite{LOGIC}, based on which we build our preference dataset. }
    \label{fig:distribution}
    \vspace{-4mm}
\end{figure}

Conventionally, after the SFT phase, $\pi_{\text{SFT}}$ is prompted with input $x$ to produce pairs of outputs $(y_1, y_2)\!\sim\!\pi_\text{SFT}(y \vert x)$, which are then presented to human annotators to rank as preferred and dispreferred responses.
Our objective is to reduce logical fallacy errors in the model's outputs, therefore, including a diverse range of fallacy types in the preference data is essential, as these may not be sufficiently represented in the model's outputs.
We define $13$ categories of logical fallacy errors (see \Figure~\ref{fig:distribution}). However, there are two key challenges: 
\textbf{(i)} determining the appropriate distribution of logical fallacy errors in the preference data, and \textbf{(ii)} automatically collecting such fallacy arguments.
To address the first concern, we leverage the LOGIC dataset \cite{LOGIC}, which was carefully curated through extensive web crawling and data collection from diverse online sources.  
This dataset reflects the distribution of fallacies in real-world scenarios, providing a realistic foundation for mitigating fallacies in everyday argumentative discourse.
This data consists of labelled logical fallacies, which we use only as examples when addressing the second concern--generating synthetic fallacious arguments given topics.
Using the \textsc{ExplaGraphs} dataset defined as $\mathcal{D}$ in \Section~\ref{pbformulation}--consisting of topics, stances and arguments--we use ChatGPT (\texttt{gpt-3.5-turbo}) to build preference pairs by generating an equivalent fallacious argument $y_l$ for each valid argument $y_w$ in $\mathcal{D}$. 
To ensure arguments generated by ChatGPT are indeed fallacies, we provide a definition of the specific fallacy being generated and examples of that fallacy type from the LOGIC dataset. 
To populate our preference dataset and ensure it spans across the most types of fallacies, we generate four fallacious arguments with different fallacy types sampled from the distribution in \Figure~\ref{fig:distribution} for each $y_w$. 
The original dataset $\mathcal{D}=\{t^{(i)},s^{(i)}, y_w^{(i)} \}^{N}_{i=1}$ containing the topic, stance and argument is now augmented with fallacies and their labels, denoted as $y_l$ and $k$ respectively. We define the preference dataset as $\mathcal{D'}=\{t^{(i)},s^{(i)}, y_w^{(i)}, y_l^{(i)}, k^{(i)} \}^{M}_{i=1}$, where we have $M$ pairs of preferred ($y_w$) and dispreferred ($y_l$) samples, with $k^{(i)}$ the fallacy type of the dispreferred argument $y_l^{(i)}$. 
The test set is not augmented with fallacies, as we use only the topics and stances at inference time to evaluate the quality and logical soundness of the arguments generated.
More details on the generations and our prompt designs for ChatGPT are presented in \Appendix~\ref{gpt-fallacies}.
Finally, our augmented fallacy argument dataset consists of the train-test split shown in \Table~\ref{tab:traintestsplit}.

\begin{table}[!htbp]
    \centering
    \resizebox{\linewidth}{!}{
    \begin{tabular}{lcc}
        \toprule
          & \textbf{\# Train} & \textbf{\# Test}   \\
        \midrule
        \textsc{ExplaGraphs} data \cite{cckg}   & 1,968 & 400 \\ 
        Generated Fallacies            & 7,872 & - \\ 
        \midrule
        Total       & 7,872 & 400\\
        \bottomrule
    \end{tabular}
    }
    \caption{Train-Test split of our preference dataset.}
    \label{tab:traintestsplit}
    \vspace{-4mm}
\end{table}


\subsection{Preference Learning Phase}\label{sec:rl_phase} 
In this work, we use four preference learning algorithms:  PPO, DPO, KTO, and CPO. Among these, only PPO requires explicit feedback from a reward model. For the other methods, we apply the preference optimization using $\pi_\text{SFT}$ as a reference model and the preference data $\mathcal{D}'$. 

\paragraph{Explicit Reward Modelling.} We use the dataset $\mathcal{D}'$ to train the Electra model \cite{electra} to learn to predict reward values.  

\paragraph{Implicit Reward Modeling.}
Methods like DPO, KTO, and CPO employ contrastive loss to derive implicit rewards from preference datasets.
Note that CPO \cite{cpo} is a reference-free method that does not require a reference policy. The different methods are detailed in \Appendix~\ref{sec:literature:preference}.



\subsection{Fallacy-Informed Preference Optimization (FIPO)}\label{sec:loss_func} 

Despite the preference optimization, 
models persistently generate specific types of logical fallacies, particularly \textit{faulty generalization} and \textit{false causality} arguments  (\Table~\ref{tab:fallacies-gpt4}). This can be attributed to the fact that the models do not explicitly learn about the fallacy types. 
Hence, we propose \ourMethod, which uses a classification head attached to the generative model to calculate a weighted cross-entropy loss for the preferred and dispreferred samples. 
Recall the preference dataset $\mathcal{D}'\!= \{t^{(i)},s^{(i)}, y_w^{(i)}, y_l^{(i)}, k^{(i)}\}_{i=1}^M$ where $k\!\in\![1, 13]$ 
is the fallacy type of $y_l$. We also label the preferred samples $y_w$ as '\textit{Not a Fallacy}' ($k=0$).
Secondly, after a forward pass through the language model $\pi_\theta$, we extract the hidden state for the last token from the last hidden layer, defined as:
\vspace{-1mm}
\begin{equation}
\label{laststate}
\mathbf{h}_\theta(y|t,s)\coloneq\pi_\theta(y|t,s)^{L,T}
    \vspace{-1mm}
\end{equation}
 where $L$ represents the total number of layers in the base model and $T$ denotes the position of the last token. This hidden state is fed into the classification head using a linear layer, and the resulting output defines the probability for fallacy type $k$:
\begin{equation}
\label{probafunction} 
 \mathbb{P}_{\mathbf{h}_\theta}^k(y|t,s) = \text{Softmax}(\mathbf{W}\mathbf{h}_\theta(y|t,s) + \mathbf{b})_k
\end{equation}
\normalsize
where $\mathbf{W}$ is the linear layer's weight matrix, and $\mathbf{b}$ is the corresponding bias term. 
To avoid penalizing the model equally for misclassifying different types of fallacies, we propose to guide the model to prioritize the most frequent fallacy types. This approach ensures that the model accurately identifies the most occurring fallacies.
We define weights $w_k$ for each fallacy type $k$ as its frequency in $\mathcal{D'}$, and $w_0$ as the minimum value of these frequencies, which is designed to let the model focus more on the fallacies in the less preferred samples rather than the non-fallacy samples during the preference optimization process: $w_k\!=\!\frac{1}{M}\sum_{i=1}^{M}\mathbbm{1}\{{k^{(i)}\!=k}\}$ and $ w_0 = \min_k w_k$
where $\mathbbm{1}$ is the indicator function. 
Using these weights, and the definitions of \Equation \ref{laststate} and \ref{probafunction}, we define the fallacy-informed classification loss as a weighted cross-entropy loss:
\begin{multline}  
\label{eq:clf_loss}
     \mathcal{L}_\text{CLF}(\pi_\theta)\!= \!-\mathbb{E}_{(t,s, y_w, y_l, k)\sim \mathcal{D'}}\\
     \left[w_0\log {\mathbb{P}}_{\mathbf{h}_\theta}^0(y_w|t,s) + w_k \log {\mathbb{P}}_{\mathbf{h}_\theta}^k(y_l|t,s) \right]
\end{multline}
The resulting loss function, termed \textbf{F}allacy \textbf{I}nformed \textbf{P}reference \textbf{O}ptimization loss, combines the loss from the preference optimization with our classification loss ($\mathcal{L}_\text{CLF}$). In our work, CPO \cite{cpo} is the method with which we combine our loss since it has the best trade-off between \wrate and \frate (see \Figure~\ref{fig:manual-win-rate-annotations} \& \ref{fig:auto-eval}, \Table~\ref{tab:fallacies-gpt4}). The resulting loss is:
\begin{equation}
    \mathcal{L}_\text{\ourMethod}(\pi_\theta) = \mathcal{L}_\text{CPO}(\pi_\theta) + \lambda  \mathcal{L}_\text{CLF}(\pi_\theta)
    \label{custom loss function equation}
\end{equation} 
where $\lambda$ is a weighting parameter to adjust the fallacy-informed loss with respect to the preference optimization loss. A more detailed description of $\mathcal{L}_\text{FIPO}$ is described in \Appendix~\ref{appendix:loss definition}. 

%% file: experimentalsetting.tex
We denote the policies obtained after the SFT phase and the alignment phase as $\pi_\text{SFT}$ and $\pi_\theta$, respectively. The policy $\pi_\theta$ is aligned using one of the following methods: PPO \cite{ppo}, DPO \cite{DPO}, 
CPO \cite{cpo}, KTO \cite{kto}, and \ourMethod.

\subsection{Datasets and Base Models}
\paragraph{Datasets.}  \label{sec:setup:dataset}
We evaluate argument generation based on the \textsc{ExplaGraphs} dataset \cite{cckg} where samples contain a \emph{Topic}, a \emph{Stance}, and short \emph{Arguments} (5-20 words), spanning a wide range of topics. We augment this dataset by generating equivalent short arguments in the form of fallacies using ChatGPT as described in \Section~\ref{sec:methodology:augmentation} and illustrated in \Figure~\ref{fig:pipeline}. The LOGIC dataset also contains short-length fallacies. Based on the length of arguments and fallacies, our study focuses only on short argumentative texts. 
The final dataset size is provided in \Table~\ref{tab:traintestsplit}.
We also perform out-of-domain analysis on a subset of samples from the Debatepedia dataset \cite{debatepedia}. 

\paragraph{Base Models.}
We use Llama-2 (7B) \cite{touvron2023llama} and Mistral (7B) \cite{mistral} as our base models.
For each alignment method, we leverage Low-Rank Adaptation (LoRA) \cite{lora}. 
This drastically reduces the number of parameters that need to be fine-tuned, from 7B to 8.3M ( $\approx\!0.12\%$). 
For both base models, we obtain a reference policy $\pi_\text{SFT}$ and an aligned policy $\pi_\theta$ for every alignment method. 
More details about hyperparameters, including training and decoding parameters, are described in \Appendix~ \ref{appendix:detailed_setup:hyperparameters}.

\subsection{Evaluation}

\label{sec:human annotation}
\paragraph{Metrics.} 
We use two metrics to evaluate the arguments generated by the aligned models compared to the baseline SFT model: the \wrate and the \textit{fallacy-rate}. The \wrate measures the proportion of instances where one argument is judged to be of higher quality than the other, while the \frate represents the proportion of logical fallacies detected in the generated arguments.

\paragraph{Human Evaluation.}
We conduct a human evaluation to compute the \textit{win-rate} to validate the relevance and quality of the generated arguments. We select 200 samples from the set of generated arguments by the aligned models, along with the corresponding topics, stances (either supporting or counter) and the equivalent arguments generated by the SFT model. Annotators perform a comparative evaluation between the SFT and aligned models by determining which argument is superior or whether both are equally good or bad. Despite the subjective nature of this task, as specific arguments may appeal differently to different individuals, we provide instructions to annotators, including selecting the argument that most clearly addresses the topic and stance. Refer to \Figure~\ref{fig:comparative_amt_page} for more details.
We also perform an annotation task where workers agree or disagree with GPT-4's fallacy predictions. This allows us to validate GPT-4 as a judge for computing the \textit{fallacy-rate}.
Annotators are recruited from Amazon Mechanical Turk ({\href{https://www.mturk.com/}{mturk.com}})
for this task. We limit our selection to native English speakers residing in the United States. The eligibility criteria for annotators include a HIT approval rate of at least 97\% and a minimum of 10,000 approved HITs.
We present more details, including the annotating instructions in \Appendix~\ref{appendix:human_annotation}. The annotators were fairly compensated. 

\begin{figure}[!h]
    \centering
    \includegraphics[width=\linewidth]{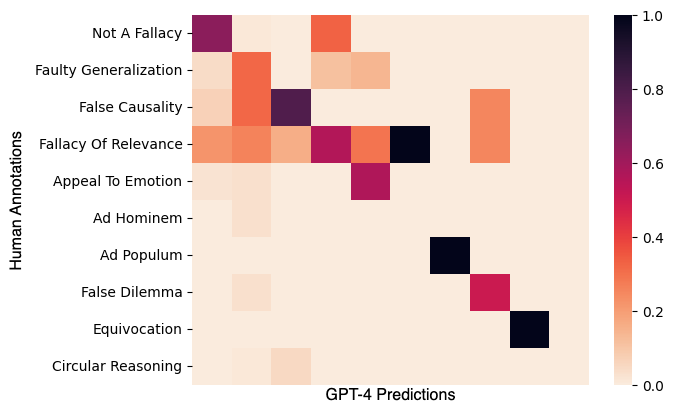}
    \caption{Heatmap for our classification compared to GPT-4's predictions. Rows are the authors'  classifications and the columns GPT-4's.}
    \vspace{-5mm}
    \label{fig:annotationheatmap}
\end{figure}

\noindent Additionally, we conduct an in-depth analysis of fallacy classification by classifying 200 arguments ourselves, aiming to provide an unbiased comparison between human evaluation and GPT-4's classifications. Since identifying logical fallacies requires domain knowledge and can be particularly challenging, our annotations helps assess GPT-4’s ability to classify and identify fallacies. This analysis sheds light on the model’s capacity to generate logical arguments. Our annotation is depicted in \Figure~\ref{fig:annotationheatmap} where the rows are our predictions and the columns are GPT-4's predictions.

\paragraph{Automatic Evaluation with GPT-4.}
We perform a pairwise comparison between the samples generated by $\pi_\text{SFT}$ and its counterparts generated by $\pi_\theta$ using the \wrate and \textit{fallacy-rate}. 
For the \textit{win-rate}, we prompt GPT-4 to decide which argument is superior, or if both are equally good. 
For the \frate, we prompt GPT-4 to evaluate the argument and detect whether the argument is any of the fallacy types out of all the ones listed in \Table~\ref{types_fallacies}, if one is present. A description of the GPT-4 evaluation and prompts is detailed in \Appendix~\ref{evalprompt}.

\paragraph{GPT-4's Reliability in Detecting Fallacies.}
\label{gpt4-reliability}
We rely on GPT-4 to compute the \textit{fallacy-rate} since  \citet{li2024reason} show GPT-4 performs well on logical fallacy detection, achieving 86\% accuracy.  
To validate GPT-4's reliability for fallacy classification, we use human annotators and ask them to \textit{agree} or \textit{disagree} with the predictions. 
We collect three annotations for each sample, consisting of ``Agree'' or ``Disagree'' responses.  
We compute the percentage agreement rate to assess the agreements between annotators and GPT-4, reflecting how often the majority vote matches GPT-4's prediction. This directly assesses GPT-4's competence in classifying fallacies, making it reliable for computing the \textit{fallacy-rate}. We also compute the agreement among annotators using Randolph's $\kappa$ \cite{randolph2005free}, which is well-suited for scenarios with rating imbalances (e.g., frequent ``Agree'' responses) by reflecting pure agreement, not assuming an expected distribution of categories. 
Randolph's $\kappa$ is computed as $\kappa=\frac{P_o-P_e}{1-P_e}$ where $P_o$ is the observed agreement: 
\vspace{-2mm}
\begin{equation}
    P_o =\frac{1}{Nn(n\!-1)}\sum^N_{i=1}\sum_{k\in\{A,D\}}n_{ik}^2 - Nn
    \vspace{-1mm}
\end{equation}
where $N$ is the number of samples annotated, $n$ is the number of annotators per sample (3 in our case) and $n_{iA}$ is the number of agreements and $n_{iD}$ the number of disagreements in each sample. $P_e=0.5$, is the expected agreement for the two categories ('Agree' and 'Disagree').
The agreement rates are displayed in \Table~\ref{tab:agreement_annotation}, and the results demonstrate the effectiveness and reliability of using GPT-4 for classifying fallacies, showing substantial agreement among annotators (0.64) and a high majority agreement ratio (0.955). More details are provided in \Appendix~\ref{appendix:human_classification}.
\begin{table}[htp!]
    \centering
    \resizebox{0.7\linewidth}{!}{
    \begin{tabular}{ll}
    \toprule
         \textbf{Agreement Metric} & \textbf{Value} \\ \hline
         Randolph's-$\kappa$& 0.640  \\
         \hline
         Majority agreement ratio & 0.955\\
         \bottomrule
    \end{tabular}
    }
    \caption{Agreement scores. Randolph's-$\kappa$ reflects agreements among annotators and majority agreement computes the agreement rate between annotators and GPT-4.}
    \label{tab:agreement_annotation}
    \vspace{-3mm}
\end{table}

Additionally, to verify GPT-4's reliability, we (the authors) perform a fallacy classification task on a set of 200 generated arguments at inference as an in-depth analysis of differences with GPT-4. The heatmap in \Figure~\ref{fig:annotationheatmap} shows the overlap between our classifications ($y$-axis) and GPT-4's ($x$-axis).
The heatmap entries are normalized to show the degree of alignment between our annotations and GPT-4’s predictions. We observe that most classifications cluster around the first few fallacy types—faulty generalization, false causality, and fallacy of relevance. The most significant disparity arises in the fallacy of relevance, where we identify this fallacy more frequently.
Additionally, the heatmap shows notable overlap between faulty generalization and false causality-when we classify an argument as false causality, GPT-4 often predicts it as a faulty generalization instead.
This suggests that the model may struggle to distinguish between broad overgeneralization and causal misattributions, likely due to subtle linguistic differences in argument structure.
Although the overall disagreement is limited, these findings highlight the difficulty of fallacy classification. Even with a solid understanding of logical fallacies, arguments can be ambiguous and complex, making misclassification a frequent challenge.

\subsection{Ablation Study}
\label{appendix:ablation study}
To validate the effectiveness of our method, we conducted two ablation studies:

 \noindent \textit{Dataset Uniformity}: The first study involves modifying the training dataset to include an equal number of samples for each fallacy type. 
 For this study, we created a uniformly distributed dataset by downsampling, resulting in a dataset comprising 2,522 samples (194 per fallacy type).

\noindent \textit{Unweighted Cross-Entropy}: The second study examines the impact of applying FIPO with unweighted cross-entropy. 
This study uses unweighted cross-entropy to demonstrate the justification for our design, as the fallacy misclassification rates are higher with unweighted cross-entropy.

Table \ref{tab:ablation} shows a sharp increase in \textit{fallacy rates}, underscoring the importance of accounting for the natural distribution of fallacy types and incorporating a weighted cross-entropy classification loss. 

\begin{table}[!h]
    \centering
    \resizebox{0.84\linewidth}{!}{
    \begin{tabular}{lc}
        \toprule
         &Fallacy Rates \\
         \midrule
          Dataset Uniformity & 37.5\% \\
          \midrule
          Unweighted Cross-Entropy &29\%\\
          \midrule
          FIPO & 17\%\\
        \bottomrule   
    \end{tabular}
    }
    \caption{Ablation study proving the effectiveness of imbalanced fallacy types and weighted cross-entropy.  }
    \label{tab:ablation}
    \vspace{-4mm}
\end{table}

%% file: results.tex
As outlined in \Section~\ref{sec:human annotation}, our evaluation of the generated arguments focuses on two primary aspects: 
(i) pairwise comparison of argument quality between the reference policy $\pi_\text{SFT}$ and the aligned policies $\pi_\theta$, which is detailed in \Section~\ref{sec:results:pairwise}; 
and (ii) the analysis of \frate across different preference optimization methods in \Section~\ref{sec:results:fallacy_rate_and_type}.

\subsection{Pairwise Comparison of Different Preference Optimization Methods}  \label{sec:results:pairwise}  
We perform a pairwise comparison to compute the \wrate between arguments generated by the SFT baselines and all the aligned models.
Each argument undergoes a manual and automatic (GPT-4-based) comparative evaluation, whose results are shown in \Figure~\ref{fig:manual-win-rate-annotations} and \Table~\ref{tab:model-win-rates}. 
For the human evaluation, each sample receives three assessments. Samples lacking majority consensus among annotators are excluded from further analysis. 
From the human annotated \wrate depicted in \Figure~\ref{fig:manual-win-rate-annotations}, we address the following research questions: 
\begin{figure}[!h]
    \centering
    \includegraphics[width=\linewidth]{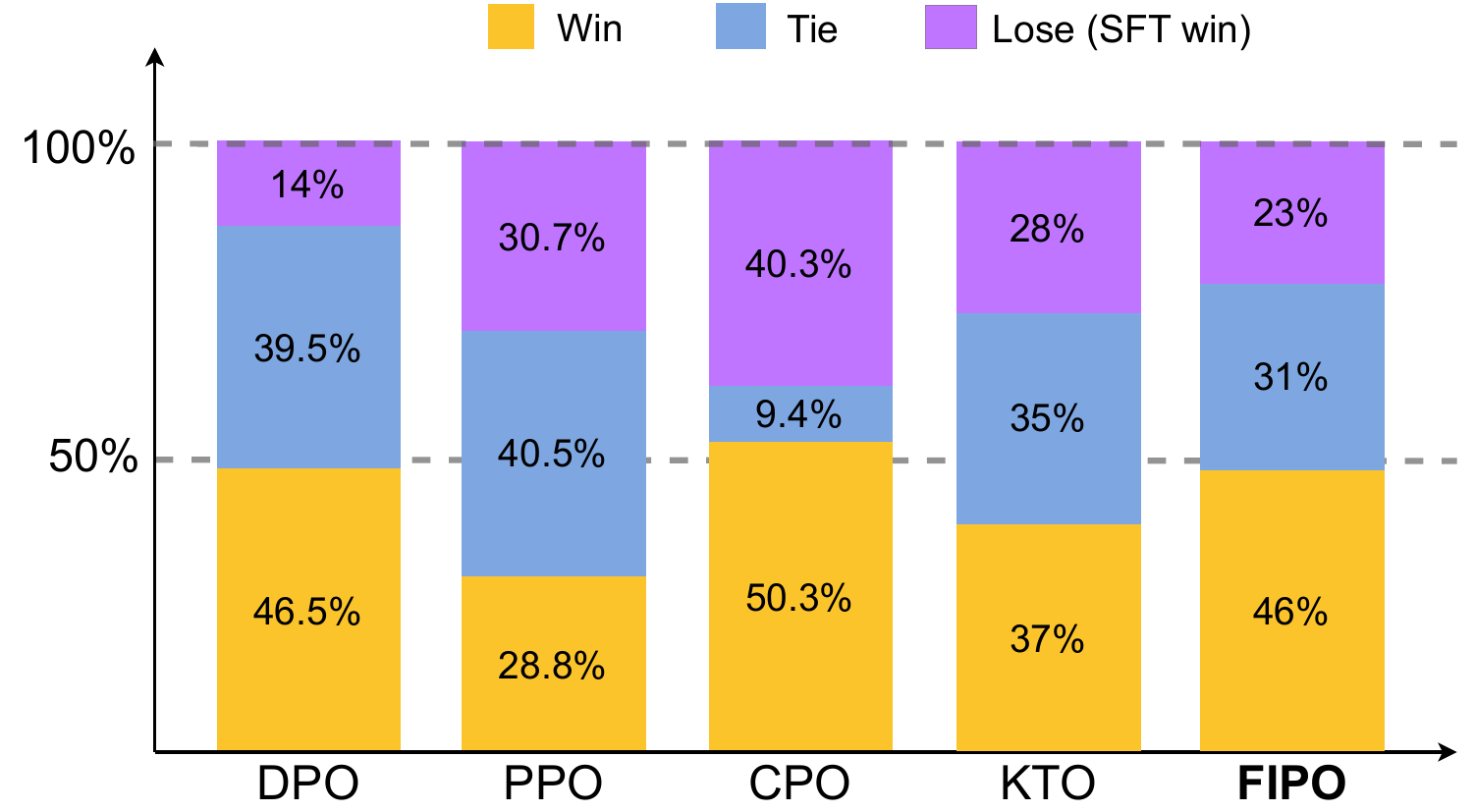}
    \caption{Human evaluation results comparing arguments generated by different preference optimization strategies using Llama-2 as the base model. The \wrate indicates how often each policy outperforms the SFT baseline regarding argument quality.}
    \label{fig:manual-win-rate-annotations}
    \vspace*{-4mm}
\end{figure}
    
\begin{figure}[!h]
    \centering
    \includegraphics[width=\linewidth]{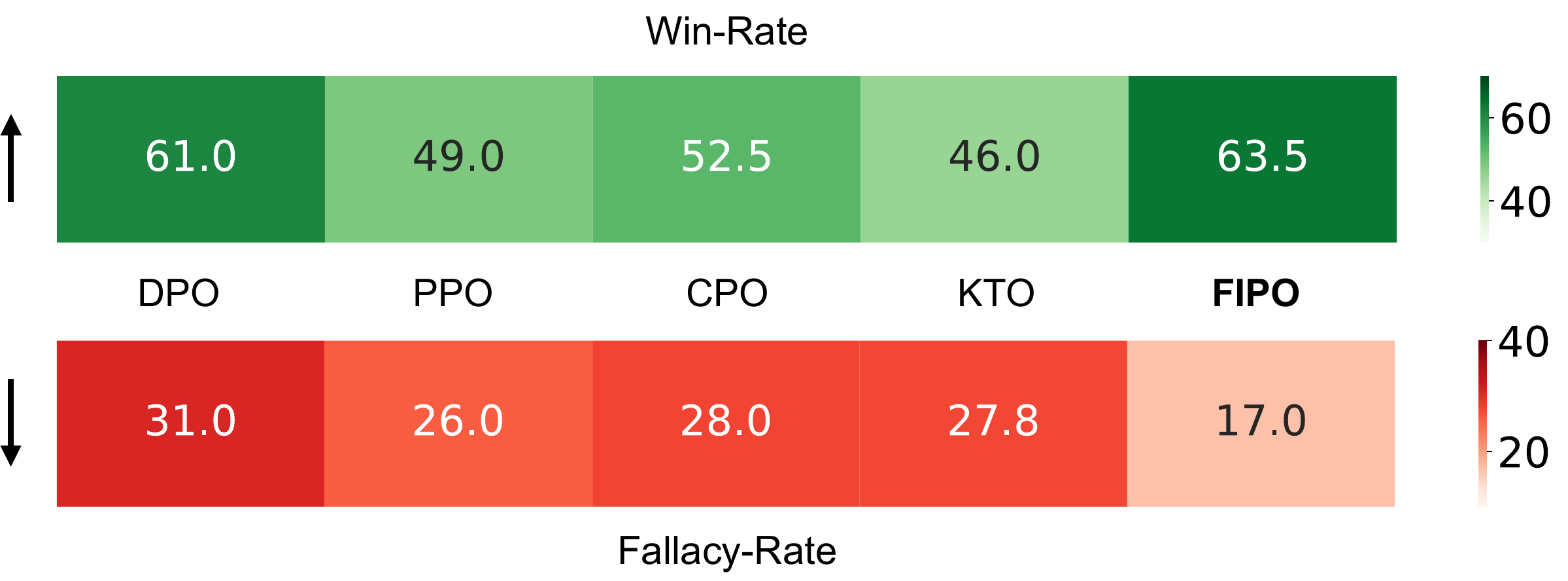}
    \caption{GPT-4 evaluation of \wrate and \textit{fallacy-rate}, for arguments generated by Llama-2. The \textit{win-rate} represents the frequency with which the aligned 
    policy outperforms SFT. The \textit{fallacy-rate} measures the proportion of the detected fallacies. The best-performing policy is our proposed \ourMethod method that achieves the lowest \textit{fallacy-rate} (17\%) and the highest \textit{win-rate} (63.5\%).}
    \label{fig:auto-eval}
\end{figure}

\noindent
\paragraph{RQ$_1$: Are preference optimization methods better than SFT?} The aligned policies outperform $\pi_\text{SFT}$ in terms of \textit{win-rate}, indicating an improvement in overall argument quality.   
DPO, CPO and \ourMethod are the only methods achieving over 40\% \textit{win-rate}, demonstrating a better ability to generate qualitative arguments. 
$\pi_\text{CPO}$ stands out as it has the highest percentage of wins (50.3\%), followed closely by $\pi_\text{DPO}$ (46.5\%). We observe similar results in the automatic \wrate evaluation with GPT-4 (\Figure~\ref{fig:auto-eval}), where DPO, CPO and FIPO have the highest \textit{win-rates}.
\noindent
\paragraph{RQ$_2$: Does FIPO improve from existing preference methods?} 
\ourMethod uses the classification loss defined in \Section~\ref{sec:loss_func} on top of CPO's loss. This is because CPO achieves the best trade-off between \wrate and \frate despite having a higher loss-rate (\Figure~\ref{fig:manual-win-rate-annotations}).
We denote this policy as $\pi_\text{\ourMethod}$. Although the \wrate of $\pi_\text{\ourMethod}$ is slightly lower at 46\% compared to CPO's 50.3\%, it is essential to note the significant decrease in loss-rate—from 40.3\% to 23\%. This reduction indicates that $\pi_\text{SFT}$ wins against $\pi_\text{\ourMethod}$~less frequently, suggesting that $\pi_\text{\ourMethod}$ produces arguments that are more qualitative compared to the baseline and to $\pi_\text{CPO}$. This improvement highlights the benefits of including more fine-grained details in FIPO's loss, making models more aware of logical fallacies, as \ourMethod~now yields outcomes that are not only equivalent but often superior to those generated by the $\pi_\text{SFT}$. 
Similar observations are drawn from the automatic \wrate evaluation in \Appendix~\ref{appendix-automated-win-rates}, where GPT-4 served as the judge to evaluate the arguments and only CPO achieved over 50\% for both Llama-2 and Mistral, as shown in \Table~\ref{tab:model-win-rates}. 

\subsection{Results for \FRate and Types} 


\begin{table*}[htbp]
    \centering
    \resizebox{\textwidth}{!}{
    \begin{tabular}{lccccccccccccc}
        \toprule
         & \multicolumn{6}{>{\columncolor{gray!20}}c}{\textbf{Llama-2 (7B)}} & &\multicolumn{6}{>{\columncolor{gray!20}}c}{\textbf{Mistral (7B)}} \\
        \cmidrule(lr){2-7} \cmidrule(lr){9-14}
        \textbf{Fallacy Types} & \textbf{SFT} & \textbf{DPO} 
        & \textbf{PPO} & \textbf{CPO} & \textbf{KTO}
        & \textbf{\ourMethod}
        & &\textbf{SFT} & \textbf{DPO}  & \textbf{PPO} & \textbf{CPO} & \textbf{KTO} 
        & \textbf{\ourMethod} \\
        \midrule
        \text{Faulty Generalization} & 27.5    & 21   & 17.5 & 19.25 & 21 
        &  7& &
        23 & 24 & 22.25 & 
        22.25 & 21 
        & 9.5\\
        \text{False Causality}      & 2.5     & 5    
        & 4.25 & 4.75  & 4.5 
        & 3.5& &
        5.25 & 5.75 &5 
        & 4& 3.5 
        &4\\
        \text{Appeal To Emotion}    & 1    & 1.25   
        & 0.75& 1.75 & -  
        & 2.5& &
        1.25 & 1.75  & 0.25 & 1.5&1.75
        & 3\\
        \text{Equivocation}         & 1  & 1  
        & 1.25  &0.25& 0.75
        & -& &
        0.75 & - & 0.5 & 0.25&0.25 
        & -\\
        \text{Fallacy of Relevance} & 0.5     & 0.25   
        & 0.75 &0.25&0.25 
        & -& &
        - & 0.5& - &0.75&0.25 
        & 0.5\\
        \text{Circular Reasoning}   & 1    & -  
        & 1.25 &-&0.75 
        &1.5&&
        0.75& 0.25&  - &-&- 
        & 0.5\\
        
        \text{Ad Populum}           & -        & 1.25    
        & -  &0.5&-  
        & -&
        & 0.25 & 0.25  & 0.25 &1&0.25 
        & 1\\
        \text{False Dilemma}        & 1       & 1.25   
        & -  &1 &0.25    
        & 2.5& &
        1 & 1  & 0.5 &0.25&0.75 
        &1\\
        \text{Ad Hominem}           & -        & -  
        & 0.25 &0.25&0.25 
        & - & &
        0.25 & 0.25 & 0.25  &-&- 
        &0.5\\
        \textbf{Not A Fallacy}    & 65.5    & 69   
        & 74 &  72& 72.25 
        & {83}&& 67.5 &66.25&71 & 70& 72.25 
        & {80.5}\\
        \midrule
        \FRate $\downarrow$     & 34.5    & 31   & \uline{26} &  28 & {27.75} 
        & \textbf{17} && 32.5  & 33.75 & 29 & 30 & \uline{27.75}  & \textbf{19.5} \\ 
        \bottomrule 
    \end{tabular}
    }
    \caption{\Frate (in percentages) of each policy, as detected by GPT-4. 
    We omit other fallacy types as none of them were reported by GPT-4. 
    \ourMethod is the top-performing method, producing the least amount of fallacies.
    }
    \label{tab:fallacies-gpt4}
\end{table*}

\label{sec:results:fallacy_rate_and_type}
Evaluating text segments to identify logical fallacies poses inherent challenges for humans. Detecting such fallacies demands an extensive understanding of logical principles and argumentative structures. Without a robust grasp of logical fallacies, differentiating between valid and flawed reasoning becomes difficult. Additionally,  personal biases and preconceptions can also cloud judgment, leading to overlooked fallacies or biased interpretations of arguments. 
We report GPT-4's evaluations in \Table~\ref{tab:fallacies-gpt4}. 
Based on the results, we address the following research questions:

\noindent\paragraph{RQ$_3$: Do preference optimization methods mitigate logical fallacy errors?}
The aligned policies produce fewer fallacies compared to the SFT baselines. Specifically, every alignment method outperforms $\pi_\text{SFT}$ for Llama-2. For Llama-2 and Mistral, DPO is the method that improves the least, and even produces more fallacies than $\pi_\text{SFT}$ with Mistral, having a higher \textit{fallacy-rate}. The other methods (PPO, CPO and KTO) consistently outperform SFT and produce fewer fallacies. 

\noindent \paragraph{RQ$_4$: Does FIPO further reduce logical fallacy errors?} The least fallacy producing policy is $\pi_\text{\ourMethod}$, achieving a \frate of 17\% for Llama-2, outperforming the previous best of 26\% (PPO). For Mistral, \ourMethod has a \frate of 19.5\%, outperforming the previous best of 27.75\% (KTO) (\Table~\ref{tab:fallacies-gpt4}). More specifically, \ourMethod, based on CPO,~beats CPO by 11\% and 10.5\% for Llama-2 and Mistral, respectively. This highlights the utility of the classification loss, indicating that the policies have a better understanding of logical fallacies than regular preference optimization.

\noindent \paragraph{RQ$_5$: What is the most observed fallacy type?} The most frequently observed fallacy produced across all policies is \textit{Faulty Generalization}. For $\pi_\text{\ourMethod}$, the occurrence for this type is only 7\%, effectively integrating the concept of generalization. An example of this occurrence is illustrated in \Table~\ref{tab:examples_fallacies}. As  \textit{Faulty Generalization} is the most frequent fallacy type in our preference dataset (18\%, \Figure~\ref{fig:distribution}), the weight assigned to this fallacy type in \Equation~\ref{eq:clf_loss} is the largest. Consequently, a higher classification loss is incurred if these fallacies are misclassified, enhancing the language model's ability to accurately identify and reduce occurrences of \textit{Faulty Generalization}. We also observe that of the 13 fallacy types, GPT-4 never classifies arguments in the following classes: \textit{Fallacy of logic, credibility, extension} and \textit{intentional}.

\subsection{Out-of-Domain Analysis}
To showcase the effectiveness of alignment methods in argument generation, we sample a test-set of 100 different topics from the Debatepedia dataset \cite{debatepedia}, and perform inference using the models previously trained on our preference dataset. Using GPT-4 as an evaluator, we compute the \textit{win-rates} and \textit{fallacy-rates}, presented in \Table~\ref{tab:model-testing}.
Using Llama-2 as the base model, results show that $\pi_\text{\ourMethod}$ is the second-best policy in terms of minimizing fallacies (55\%), slightly behind $\pi_\text{KTO}$ (56\%).
We also find that \ourMethod~achieves the highest \textit{win-rate}, winning 62\% of the times against $\pi_\text{SFT}$. Interestingly, our results reveal that FIPO helps to reduce \textit{False Causality} and \textit{Fallacy of Relevance} fallacies. 
\begin{table}[!h]
    \centering
    \resizebox{\linewidth}{!}{
    \begin{tabular}{lcccccc}
        \toprule
        \textbf{Fallacy Types}  & \textbf{SFT} & \textbf{DPO}  &  \textbf{PPO} & \textbf{CPO} &\textbf{KTO} & \textbf{\ourMethod~} \\
        \midrule
        Faulty Generalization      & 17 & 20  & 17 & 24&17 & 18\\ 
        False Causality            & 9 &   7  & 6 & 8& 8 & 5\\ 
        Appeal To Emotion          & 7 & 16 & 13 & 13& 7 &12\\
        Fallacy of Relevance       & 12 & 6  & 6 &10 & 3& -\\ 
        Ad Populum                 & 3 & 4  & 2 &1&1& -\\ 
        False Dilemma              & 6 & 6 & 6 & 4 & 6&7\\ 
        Equivocation               & - & - & 2& 1 &1 & 2\\ 
        Circular Reasoning         & 4 & 2 & - &2&1&1\\ 
        \midrule
        \FRate $\downarrow$            & 58 & 61 & 52 & 63 & \textbf{44} & \uline{45}\\
        
        \midrule
        Win-Rate vs. SFT $\uparrow$          & - & \uline{59} & 54 & 43 & 55 & \textbf{62} \\

        \bottomrule
    \end{tabular}
    }
    \caption{Fallacies generated by different alignment methods in the out-of-domain setting, detected by GPT-4.  
    We omit the other fallacy types, as none of them were reported as such by GPT-4. We also evaluate the \wrate and observe that \ourMethod achieves the highest one and is the second-best policy at not generating fallacies.}
    \label{tab:model-testing}
\end{table}

%% file: conclusion.tex
In this work, we investigate the impact of logical fallacies on argument generation and introduce FIPO, a novel framework designed to improve the logical soundness of arguments by including a classification loss during the preference optimization phase. Both human and automatic evaluations show that our method produces higher-quality arguments and achieves lower \textit{fallacy-rates}. These findings underscore the importance of addressing logical fallacies in improving argument generation.

%% file: acl2023.bbl
\begin{thebibliography}{45}
\expandafter\ifx\csname natexlab\endcsname\relax\def\natexlab#1{#1}\fi

\bibitem[{Aristotle(2006)}]{aristotle}
Aristotle. 2006.
\newblock \emph{On Sophistical Refutations}.
\newblock The Internet Classics Archive.

\bibitem[{Bird and Loper(2004)}]{bird-loper-2004-nltk}
Steven Bird and Edward Loper. 2004.
\newblock \href {https://aclanthology.org/P04-3031} {{NLTK}: The natural language toolkit}.
\newblock In \emph{Proceedings of the {ACL} Interactive Poster and Demonstration Sessions}, pages 214--217, Barcelona, Spain. Association for Computational Linguistics.

\bibitem[{Cabrio and Villata(2012)}]{debatepedia}
Elena Cabrio and Serena Villata. 2012.
\newblock \href {https://doi.org/10.3233/978-1-61499-098-7-205} {Natural language arguments: A combined approach.}
\newblock \emph{Frontiers in Artificial Intelligence and Applications}, 242.

\bibitem[{Chen et~al.(2023{\natexlab{a}})Chen, Cheng, Luu, and Bing}]{Chen2023ExploringTP}
Guizhen Chen, Liying Cheng, Anh~Tuan Luu, and Lidong Bing. 2023{\natexlab{a}}.
\newblock \href {https://api.semanticscholar.org/CorpusID:265213435} {Exploring the potential of large language models in computational argumentation}.
\newblock \emph{ArXiv}, abs/2311.09022.

\bibitem[{Chen et~al.(2023{\natexlab{b}})Chen, Ma, Song, Cao, Zhang, and Li}]{chen2023learning}
Meiqi Chen, Yubo Ma, Kaitao Song, Yixin Cao, Yan Zhang, and Dongsheng Li. 2023{\natexlab{b}}.
\newblock Learning to teach large language models logical reasoning.
\newblock \emph{arXiv preprint arXiv:2310.09158}.

\bibitem[{Chia et~al.(2022)Chia, Bing, Poria, and Si}]{chia-etal-2022-relationprompt}
Yew~Ken Chia, Lidong Bing, Soujanya Poria, and Luo Si. 2022.
\newblock \href {https://doi.org/10.18653/v1/2022.findings-acl.5} {{R}elation{P}rompt: Leveraging prompts to generate synthetic data for zero-shot relation triplet extraction}.
\newblock In \emph{Findings of the Association for Computational Linguistics: ACL 2022}, pages 45--57, Dublin, Ireland. Association for Computational Linguistics.

\bibitem[{Clark et~al.(2020)Clark, Luong, Le, and Manning}]{electra}
Kevin Clark, Minh{-}Thang Luong, Quoc~V. Le, and Christopher~D. Manning. 2020.
\newblock \href {http://arxiv.org/abs/2003.10555} {{ELECTRA:} pre-training text encoders as discriminators rather than generators}.
\newblock \emph{CoRR}, abs/2003.10555.

\bibitem[{Ethayarajh et~al.(2024)Ethayarajh, Xu, Muennighoff, Jurafsky, and Kiela}]{kto}
Kawin Ethayarajh, Winnie Xu, Niklas Muennighoff, Dan Jurafsky, and Douwe Kiela. 2024.
\newblock \href {http://arxiv.org/abs/2402.01306} {Kto: Model alignment as prospect theoretic optimization}.

\bibitem[{Evans(2002)}]{humanreasoning}
Jonathan St~BT Evans. 2002.
\newblock Logic and human reasoning: an assessment of the deduction paradigm.
\newblock \emph{Psychological bulletin}, 128(6):978.

\bibitem[{Hu et~al.(2021)Hu, Shen, Wallis, Allen-Zhu, Li, Wang, Wang, and Chen}]{lora}
Edward~J Hu, Yelong Shen, Phillip Wallis, Zeyuan Allen-Zhu, Yuanzhi Li, Shean Wang, Lu~Wang, and Weizhu Chen. 2021.
\newblock Lora: Low-rank adaptation of large language models.
\newblock \emph{arXiv preprint arXiv:2106.09685}.

\bibitem[{Hua et~al.(2019)Hua, Hu, and Wang}]{hua-etal-2019-argument-generation}
Xinyu Hua, Zhe Hu, and Lu~Wang. 2019.
\newblock \href {https://doi.org/10.18653/v1/P19-1255} {Argument generation with retrieval, planning, and realization}.
\newblock In \emph{Proceedings of the 57th Annual Meeting of the Association for Computational Linguistics}, pages 2661--2672, Florence, Italy. Association for Computational Linguistics.

\bibitem[{Hua and Wang(2018)}]{hua-wang-2018-neural}
Xinyu Hua and Lu~Wang. 2018.
\newblock \href {https://doi.org/10.18653/v1/P18-1021} {Neural argument generation augmented with externally retrieved evidence}.
\newblock In \emph{Proceedings of the 56th Annual Meeting of the Association for Computational Linguistics (Volume 1: Long Papers)}, pages 219--230, Melbourne, Australia. Association for Computational Linguistics.

\bibitem[{Jiang et~al.(2023)Jiang, Sablayrolles, Mensch, Bamford, Chaplot, Casas, Bressand, Lengyel, Lample, Saulnier et~al.}]{mistral}
Albert~Q Jiang, Alexandre Sablayrolles, Arthur Mensch, Chris Bamford, Devendra~Singh Chaplot, Diego de~las Casas, Florian Bressand, Gianna Lengyel, Guillaume Lample, Lucile Saulnier, et~al. 2023.
\newblock Mistral 7b.
\newblock \emph{arXiv preprint arXiv:2310.06825}.

\bibitem[{Jin et~al.(2022)Jin, Lalwani, Vaidhya, Shen, Ding, Lyu, Sachan, Mihalcea, and Schoelkopf}]{LOGIC}
Zhijing Jin, Abhinav Lalwani, Tejas Vaidhya, Xiaoyu Shen, Yiwen Ding, Zhiheng Lyu, Mrinmaya Sachan, Rada Mihalcea, and Bernhard Schoelkopf. 2022.
\newblock \href {https://doi.org/10.18653/v1/2022.findings-emnlp.532} {Logical fallacy detection}.
\newblock In \emph{Findings of the Association for Computational Linguistics: EMNLP 2022}, pages 7180--7198, Abu Dhabi, United Arab Emirates. Association for Computational Linguistics.

\bibitem[{Karpukhin et~al.(2020)Karpukhin, Oguz, Min, Lewis, Wu, Edunov, Chen, and Yih}]{wiki-dpr}
Vladimir Karpukhin, Barlas Oguz, Sewon Min, Patrick Lewis, Ledell Wu, Sergey Edunov, Danqi Chen, and Wen-tau Yih. 2020.
\newblock \href {https://doi.org/10.18653/v1/2020.emnlp-main.550} {Dense passage retrieval for open-domain question answering}.
\newblock In \emph{Proceedings of the 2020 Conference on Empirical Methods in Natural Language Processing (EMNLP)}, pages 6769--6781, Online. Association for Computational Linguistics.

\bibitem[{Lee et~al.(2023)Lee, Phatale, Mansoor, Lu, Mesnard, Bishop, Carbune, and Rastogi}]{lee2023rlaif}
Harrison Lee, Samrat Phatale, Hassan Mansoor, Kellie Lu, Thomas Mesnard, Colton Bishop, Victor Carbune, and Abhinav Rastogi. 2023.
\newblock Rlaif: Scaling reinforcement learning from human feedback with ai feedback.
\newblock \emph{arXiv preprint arXiv:2309.00267}.

\bibitem[{Lewis et~al.(2020)Lewis, Perez, Piktus, Petroni, Karpukhin, Goyal, K{\"u}ttler, Lewis, Yih, Rockt{\"a}schel et~al.}]{rag}
Patrick Lewis, Ethan Perez, Aleksandra Piktus, Fabio Petroni, Vladimir Karpukhin, Naman Goyal, Heinrich K{\"u}ttler, Mike Lewis, Wen-tau Yih, Tim Rockt{\"a}schel, et~al. 2020.
\newblock Retrieval-augmented generation for knowledge-intensive nlp tasks.
\newblock \emph{Advances in Neural Information Processing Systems}, 33:9459--9474.

\bibitem[{Li et~al.(2024)Li, Wang, Liang, Jiang, He, Xiao, and Yang}]{li2024reason}
Yanda Li, Dixuan Wang, Jiaqing Liang, Guochao Jiang, Qianyu He, Yanghua Xiao, and Deqing Yang. 2024.
\newblock Reason from fallacy: Enhancing large language models' logical reasoning through logical fallacy understanding.
\newblock \emph{arXiv preprint arXiv:2404.04293}.

\bibitem[{Lin(2004)}]{lin-2004-rouge}
Chin-Yew Lin. 2004.
\newblock \href {https://aclanthology.org/W04-1013} {{ROUGE}: A package for automatic evaluation of summaries}.
\newblock In \emph{Text Summarization Branches Out}, pages 74--81, Barcelona, Spain. Association for Computational Linguistics.

\bibitem[{Liu et~al.(2023)Liu, Iter, Xu, Wang, Xu, and Zhu}]{liu-etal-2023-g}
Yang Liu, Dan Iter, Yichong Xu, Shuohang Wang, Ruochen Xu, and Chenguang Zhu. 2023.
\newblock \href {https://doi.org/10.18653/v1/2023.emnlp-main.153} {{G}-eval: {NLG} evaluation using gpt-4 with better human alignment}.
\newblock In \emph{Proceedings of the 2023 Conference on Empirical Methods in Natural Language Processing}, pages 2511--2522, Singapore. Association for Computational Linguistics.

\bibitem[{Nakpih and Santini(2020)}]{nakpih2020automated}
Callistus~Ireneous Nakpih and Simone Santini. 2020.
\newblock Automated discovery of logical fallacies in legal argumentation.
\newblock \emph{International Journal of Artificial Intelligence and Applications (IJAIA)}, 11.

\bibitem[{Papineni et~al.(2002)Papineni, Roukos, Ward, and Zhu}]{papineni-etal-2002-bleu}
Kishore Papineni, Salim Roukos, Todd Ward, and Wei-Jing Zhu. 2002.
\newblock \href {https://doi.org/10.3115/1073083.1073135} {{B}leu: a method for automatic evaluation of machine translation}.
\newblock In \emph{Proceedings of the 40th Annual Meeting of the Association for Computational Linguistics}, pages 311--318, Philadelphia, Pennsylvania, USA. Association for Computational Linguistics.

\bibitem[{Paszke et~al.(2019)Paszke, Gross, Massa, Lerer, Bradbury, Chanan, Killeen, Lin, Gimelshein, Antiga, Desmaison, K\"{o}pf, Yang, DeVito, Raison, Tejani, Chilamkurthy, Steiner, Fang, Bai, and Chintala}]{paszke-etal-2019-pytorch}
Adam Paszke, Sam Gross, Francisco Massa, Adam Lerer, James Bradbury, Gregory Chanan, Trevor Killeen, Zeming Lin, Natalia Gimelshein, Luca Antiga, Alban Desmaison, Andreas K\"{o}pf, Edward Yang, Zach DeVito, Martin Raison, Alykhan Tejani, Sasank Chilamkurthy, Benoit Steiner, Lu~Fang, Junjie Bai, and Soumith Chintala. 2019.
\newblock \emph{PyTorch: an imperative style, high-performance deep learning library}. Curran Associates Inc., Red Hook, NY, USA.

\bibitem[{Pedregosa et~al.(2011)Pedregosa, Varoquaux, Gramfort, Michel, Thirion, Grisel, Blondel, Prettenhofer, Weiss, Dubourg et~al.}]{pedregosa-etal-2011-scikit-learn}
Fabian Pedregosa, Ga{\"e}l Varoquaux, Alexandre Gramfort, Vincent Michel, Bertrand Thirion, Olivier Grisel, Mathieu Blondel, Peter Prettenhofer, Ron Weiss, Vincent Dubourg, et~al. 2011.
\newblock Scikit-learn: Machine learning in python.
\newblock \emph{Journal of Machine Learning Research}, 12(Oct):2825--2830.

\bibitem[{Peng et~al.(2023)Peng, Li, He, Galley, and Gao}]{peng2023instruction}
Baolin Peng, Chunyuan Li, Pengcheng He, Michel Galley, and Jianfeng Gao. 2023.
\newblock \href {http://arxiv.org/abs/2304.03277} {Instruction tuning with gpt-4}.

\bibitem[{Rafailov et~al.(2023)Rafailov, Sharma, Mitchell, Manning, Ermon, and Finn}]{DPO}
Rafael Rafailov, Archit Sharma, Eric Mitchell, Christopher~D Manning, Stefano Ermon, and Chelsea Finn. 2023.
\newblock \href {https://arxiv.org/abs/2305.18290} {Direct preference optimization: Your language model is secretly a reward model}.
\newblock In \emph{Thirty-seventh Conference on Neural Information Processing Systems}.

\bibitem[{Russell(2013)}]{Russell2013}
Bertrand Russell. 2013.
\newblock \emph{History of Western Philosophy: Collectors Edition}.
\newblock Routledge.

\bibitem[{Saha and Srihari(2023)}]{saha-srihari-2023-argu}
Sougata Saha and Rohini Srihari. 2023.
\newblock \href {https://doi.org/10.18653/v1/2023.acl-long.466} {{A}rg{U}: A controllable factual argument generator}.
\newblock In \emph{Proceedings of the 61st Annual Meeting of the Association for Computational Linguistics (Volume 1: Long Papers)}, pages 8373--8388, Toronto, Canada. Association for Computational Linguistics.

\bibitem[{Saha et~al.(2021)Saha, Yadav, Bauer, and Bansal}]{cckg}
Swarnadeep Saha, Prateek Yadav, Lisa Bauer, and Mohit Bansal. 2021.
\newblock \href {https://doi.org/10.18653/v1/2021.emnlp-main.609} {{E}xpla{G}raphs: An explanation graph generation task for structured commonsense reasoning}.
\newblock In \emph{Proceedings of the 2021 Conference on Empirical Methods in Natural Language Processing}, pages 7716--7740, Online and Punta Cana, Dominican Republic. Association for Computational Linguistics.

\bibitem[{Schick and Sch{\"u}tze(2021)}]{schick-schutze-2021-generating}
Timo Schick and Hinrich Sch{\"u}tze. 2021.
\newblock \href {https://doi.org/10.18653/v1/2021.emnlp-main.555} {Generating datasets with pretrained language models}.
\newblock In \emph{Proceedings of the 2021 Conference on Empirical Methods in Natural Language Processing}, pages 6943--6951, Online and Punta Cana, Dominican Republic. Association for Computational Linguistics.

\bibitem[{Schiller et~al.(2021)Schiller, Daxenberger, and Gurevych}]{schiller-etal-2021-aspect}
Benjamin Schiller, Johannes Daxenberger, and Iryna Gurevych. 2021.
\newblock \href {https://doi.org/10.18653/v1/2021.naacl-main.34} {Aspect-controlled neural argument generation}.
\newblock In \emph{Proceedings of the 2021 Conference of the North American Chapter of the Association for Computational Linguistics: Human Language Technologies}, pages 380--396, Online. Association for Computational Linguistics.

\bibitem[{Schulman et~al.(2017)Schulman, Wolski, Dhariwal, Radford, and Klimov}]{ppo}
John Schulman, Filip Wolski, Prafulla Dhariwal, Alec Radford, and Oleg Klimov. 2017.
\newblock \href {http://arxiv.org/abs/1707.06347} {Proximal policy optimization algorithms}.

\bibitem[{Shao et~al.(2023)Shao, Gong, Shen, Huang, Duan, and Chen}]{10.5555/3618408.3619681}
Zhihong Shao, Yeyun Gong, Yelong Shen, Minlie Huang, Nan Duan, and Weizhu Chen. 2023.
\newblock Synthetic prompting: generating chain-of-thought demonstrations for large language models.
\newblock In \emph{Proceedings of the 40th International Conference on Machine Learning}, ICML'23. JMLR.org.

\bibitem[{Sheng et~al.(2020)Sheng, Chang, Natarajan, and Peng}]{ethics2}
Emily Sheng, Kai-Wei Chang, Prem Natarajan, and Nanyun Peng. 2020.
\newblock \href {https://doi.org/10.18653/v1/2020.findings-emnlp.291} {Towards {C}ontrollable {B}iases in {L}anguage {G}eneration}.
\newblock In \emph{Findings of the Association for Computational Linguistics: EMNLP 2020}, pages 3239--3254, Online. Association for Computational Linguistics.

\bibitem[{Sourati et~al.(2023)Sourati, Ilievski, Sandlin, and Mermoud}]{CBR}
Zhivar Sourati, Filip Ilievski, H{\^o}ng-{\^A}n Sandlin, and Alain Mermoud. 2023.
\newblock Case-based reasoning with language models for classification of logical fallacies.
\newblock \emph{arXiv preprint arXiv:2301.11879}.

\bibitem[{Sun et~al.(2023)Sun, Shen, Zhou, Zhang, Chen, Cox, Yang, and Gan}]{sun2023principledriven}
Zhiqing Sun, Yikang Shen, Qinhong Zhou, Hongxin Zhang, Zhenfang Chen, David~Daniel Cox, Yiming Yang, and Chuang Gan. 2023.
\newblock \href {https://openreview.net/forum?id=p40XRfBX96} {Principle-driven self-alignment of language models from scratch with minimal human supervision}.
\newblock In \emph{Thirty-seventh Conference on Neural Information Processing Systems}.

\bibitem[{Tindale(2007)}]{tindale2007fallacies}
Christopher~W Tindale. 2007.
\newblock \emph{Fallacies and argument appraisal}.
\newblock Cambridge University Press.

\bibitem[{Christiano et~al.(2017)Christiano, Leike, Brown, Martic, Legg, and Amodei}]{christiano2017deep_rl}
Paul F. Christiano, Jan Leike, Tom Brown, Miljan Martic, Shane Legg, and Dario Amodei. 2017.  
\newblock Deep reinforcement learning from human preferences.  
\newblock \emph{Advances in Neural Information Processing Systems}, 30.

\bibitem[{Ziegler et~al.(2019)Ziegler, Stiennon, Wu, Brown, Radford, Amodei, Christiano, and Irving}]{ziegler2019fine}
Daniel M. Ziegler, Nisan Stiennon, Jeffrey Wu, Tom B. Brown, Alec Radford, Dario Amodei, Paul Christiano, and Geoffrey Irving. 2019.  
\newblock Fine-tuning language models from human preferences.  
\newblock \emph{arXiv preprint arXiv:1909.08593}.

\bibitem[{Ruiz-Dolz and Lawrence(2023)}]{ruiz-dolz-lawrence-2023-detecting}
Ramon Ruiz-Dolz and John Lawrence. 2023.  
\newblock Detecting argumentative fallacies in the wild: Problems and limitations of large language models.  
\newblock In \emph{Proceedings of the 10th Workshop on Argument Mining}, pages 1--10, Singapore. Association for Computational Linguistics.  
\newblock \url{https://aclanthology.org/2023.argmining-1.1/}.

\bibitem[{Randolph(2005)}]{randolph2005free}
Justus J. Randolph. 2005.  
\newblock Free-marginal multirater kappa (multirater $K_\text{free}$): An alternative to Fleiss' fixed-marginal multirater kappa.  
\newblock \emph{Advances in Data Analysis and Classification}.


\bibitem[{Touvron et~al.(2023)Touvron, Lavril, Izacard, Martinet, Lachaux, Lacroix, Rozi{\`e}re, Goyal, Hambro, Azhar et~al.}]{touvron2023llama}
Hugo Touvron, Thibaut Lavril, Gautier Izacard, Xavier Martinet, Marie-Anne Lachaux, Timoth{\'e}e Lacroix, Baptiste Rozi{\`e}re, Naman Goyal, Eric Hambro, Faisal Azhar, et~al. 2023.
\newblock Llama: Open and efficient foundation language models.
\newblock \emph{arXiv preprint arXiv:2302.13971}.

\bibitem[{Wachsmuth et~al.(2017)Wachsmuth, Naderi, Habernal, Hou, Hirst, Gurevych, and Stein}]{wachsmuth-etal-2017-argumentation}
Henning Wachsmuth, Nona Naderi, Ivan Habernal, Yufang Hou, Graeme Hirst, Iryna Gurevych, and Benno Stein. 2017.
\newblock \href {https://doi.org/10.18653/v1/P17-2039} {Argumentation quality assessment: Theory vs. practice}.
\newblock In \emph{Proceedings of the 55th Annual Meeting of the Association for Computational Linguistics (Volume 2: Short Papers)}, pages 250--255, Vancouver, Canada. Association for Computational Linguistics.

\bibitem[{Walton et~al.(2008)Walton, Reed, and Macagno}]{walton2008argumentation}
Douglas Walton, Christopher Reed, and Fabrizio Macagno. 2008.
\newblock \emph{Argumentation schemes}.
\newblock Cambridge University Press.

\bibitem[{Wang et~al.(2023)Wang, Kordi, Mishra, Liu, Smith, Khashabi, and Hajishirzi}]{wang-etal-2023-self-instruct}
Yizhong Wang, Yeganeh Kordi, Swaroop Mishra, Alisa Liu, Noah~A. Smith, Daniel Khashabi, and Hannaneh Hajishirzi. 2023.
\newblock \href {https://doi.org/10.18653/v1/2023.acl-long.754} {Self-instruct: Aligning language models with self-generated instructions}.
\newblock In \emph{Proceedings of the 61st Annual Meeting of the Association for Computational Linguistics (Volume 1: Long Papers)}, pages 13484--13508, Toronto, Canada. Association for Computational Linguistics.

\bibitem[{Weidinger et~al.(2021)Weidinger, Mellor, Rauh, Griffin, Uesato, Huang, Cheng, Glaese, Balle, Kasirzadeh et~al.}]{ethics1}
Laura Weidinger, John Mellor, Maribeth Rauh, Conor Griffin, Jonathan Uesato, Po-Sen Huang, Myra Cheng, Mia Glaese, Borja Balle, Atoosa Kasirzadeh, et~al. 2021.
\newblock Ethical and social risks of harm from language models.
\newblock \emph{arXiv preprint arXiv:2112.04359}.

\bibitem[{Wolf et~al.(2020)Wolf, Debut, Sanh, Chaumond, Delangue, Moi, Cistac, Rault, Louf, Funtowicz, Davison, Shleifer, von Platen, Ma, Jernite, Plu, Xu, Le~Scao, Gugger, Drame, Lhoest, and Rush}]{wolf-etal-2020-transformers}
Thomas Wolf, Lysandre Debut, Victor Sanh, Julien Chaumond, Clement Delangue, Anthony Moi, Pierric Cistac, Tim Rault, Remi Louf, Morgan Funtowicz, Joe Davison, Sam Shleifer, Patrick von Platen, Clara Ma, Yacine Jernite, Julien Plu, Canwen Xu, Teven Le~Scao, Sylvain Gugger, Mariama Drame, Quentin Lhoest, and Alexander Rush. 2020.
\newblock \href {https://doi.org/10.18653/v1/2020.emnlp-demos.6} {Transformers: State-of-the-art natural language processing}.
\newblock In \emph{Proceedings of the 2020 Conference on Empirical Methods in Natural Language Processing: System Demonstrations}, pages 38--45, Online. Association for Computational Linguistics.

\bibitem[{Xu et~al.(2024)Xu, Sharaf, Chen, Tan, Shen, Durme, Murray, and Kim}]{cpo}
Haoran Xu, Amr Sharaf, Yunmo Chen, Weiting Tan, Lingfeng Shen, Benjamin~Van Durme, Kenton Murray, and Young~Jin Kim. 2024.
\newblock \href {http://arxiv.org/abs/2401.08417} {Contrastive preference optimization: Pushing the boundaries of llm performance in machine translation}.

\bibitem[{Xu et~al.(2023)Xu, Xie, Qin, Tao, and Wang}]{peft}
Lingling Xu, Haoran Xie, Si-Zhao~Joe Qin, Xiaohui Tao, and Fu~Lee Wang. 2023.
\newblock Parameter-efficient fine-tuning methods for pretrained language models: A critical review and assessment.
\newblock \emph{arXiv preprint arXiv:2312.12148}.

\end{thebibliography}
